\documentclass[12pt]{article}

\usepackage{scicite}

\usepackage{bm}
\usepackage{amsfonts}
\usepackage{amssymb}
\usepackage{amsmath} %
\usepackage{xr}
\usepackage{hyperref}
\usepackage{subfigure}
\usepackage{float}
\usepackage{graphicx}
\usepackage{xspace}
\usepackage{footmisc}
\usepackage[font=small,labelfont=bf]{caption}
\usepackage{tabularx, makecell, booktabs}
\usepackage{color}
\usepackage[capitalize]{cleveref}
\usepackage{times}

\usepackage{comment}
\usepackage[binary-units]{siunitx}
\usepackage{relsize}
\usepackage{ifthen}
\usepackage[colorinlistoftodos]{todonotes}

\usepackage[vlined,ruled,linesnumbered]{algorithm2e}
\usepackage{graphics} 
\usepackage{rotating}
\usepackage{color}
\usepackage{enumerate}
\usepackage[T1]{fontenc}
\usepackage{psfrag}
\usepackage{epsfig} 
\usepackage{booktabs}
\usepackage{graphicx,url}
\usepackage{multirow}
\usepackage{array}
\usepackage{latexsym}
\usepackage{amsfonts}
\usepackage{amsmath}
\usepackage{amssymb}
\usepackage{xstring}
\usepackage[noend]{algorithmic}
\usepackage{multirow}
\usepackage{xcolor}
\usepackage{prettyref}
\usepackage{flexisym}
\usepackage{bigdelim}
\usepackage{breqn} 
\usepackage{listings}
\let\labelindent\relax
\usepackage{enumitem}
\usepackage{xspace}
\usepackage{bm}
\graphicspath{{./figures/}}
\usepackage{tikz}
\usetikzlibrary{matrix,calc}

\usepackage{mdwlist}

\let\itemize\undefined
\makecompactlist{itemize}{stditemize}

\newrefformat{prob}{Problem\,\ref{#1}}
\newrefformat{def}{Definition\,\ref{#1}}
\newrefformat{sec}{Section\,\ref{#1}}
\newrefformat{sub}{Section\,\ref{#1}}
\newrefformat{prop}{Proposition\,\ref{#1}}
\newrefformat{app}{Appendix\,\ref{#1}}
\newrefformat{alg}{Algorithm\,\ref{#1}}
\newrefformat{cor}{Corollary\,\ref{#1}}
\newrefformat{thm}{Theorem\,\ref{#1}}
\newrefformat{lem}{Lemma\,\ref{#1}}
\newrefformat{fig}{Fig.\,\ref{#1}}
\newrefformat{tab}{Table\,\ref{#1}}

\newcommand{\bdmath}{\begin{dmath}}
\newcommand{\edmath}{\end{dmath}}
\newcommand{\beq}{\begin{equation}}
\newcommand{\eeq}{\end{equation}}
\newcommand{\bdm}{\begin{displaymath}}
\newcommand{\edm}{\end{displaymath}}
\newcommand{\bea}{\begin{eqnarray}}
\newcommand{\eea}{\end{eqnarray}}
\newcommand{\beal}{\beq \begin{array}{ll}}
\newcommand{\eeal}{\end{array} \eeq}
\newcommand{\beas}{\begin{eqnarray*}}
\newcommand{\eeas}{\end{eqnarray*}}
\newcommand{\ba}{\begin{array}}
\newcommand{\ea}{\end{array}}
\newcommand{\bit}{\begin{itemize}}
\newcommand{\eit}{\end{itemize}}
\newcommand{\ben}{\begin{enumerate}}
\newcommand{\een}{\end{enumerate}}

\newcommand{\eg}{\emph{e.g.,}\xspace}
\newcommand{\ie}{\emph{i.e.,}\xspace}

\newcommand{\M}[1]{{\bm #1}} 
\renewcommand{\boldsymbol}[1]{{\bm #1}}

\newcommand{\hide}[1]{}

\newcommand{\hiddenText}{{\color{gray} hidden text.}}
\newcommand{\hideWithText}[1]{\hiddenText}

\newcommand{\tran}{^{\mathsf{T}}}

\newcommand{\Real}[1]{ { {\mathbb R}^{#1} } }

\newcommand{\SOthree}{\ensuremath{\mathrm{SO}(3)}\xspace}

\newcommand{\MJ}{\M{J}}

\newcommand{\MR}{\M{R}}

\newcommand{\vb}{\boldsymbol{b}}

\newcommand{\ve}{\boldsymbol{e}}

\newcommand{\vg}{\boldsymbol{g}}

\newcommand{\vp}{\boldsymbol{p}}

\newcommand{\vtheta}{\boldsymbol{\theta}}

\newcommand{\vtau}{\boldsymbol{\tau}}

\newcommand{\blue}[1]{{\color{blue}#1}}

\newcommand{\linkToPdf}[1]{\href{#1}{\blue{(pdf)}}}
\newcommand{\linkToPpt}[1]{\href{#1}{\blue{(ppt)}}}
\newcommand{\linkToCode}[1]{\href{#1}{\blue{(code)}}}
\newcommand{\linkToWeb}[1]{\href{#1}{\blue{(web)}}}
\newcommand{\linkToVideo}[1]{\href{#1}{\blue{(video)}}}
\newcommand{\linkToMedia}[1]{\href{#1}{\blue{(media)}}}
\newcommand{\award}[1]{\xspace} 

\newenvironment{sciabstract}{%
\begin{quote} \bf}
{\end{quote}}

\title{Aggressive Aerial Grasping using a Soft Drone \\ with Onboard Perception} 

\author
{Samuel Ubellacker$^{1\ast}$, Aaron Ray$^{1}$, James M. Bern$^{2}$, Jared Strader$^{1}$, Luca Carlone$^{1}$\\
\\
\normalsize{$^{1}$Massachusetts Institute of Technology}\\
\normalsize{$^{2}$Williams College}\\
 \vspace{-5mm}
 \\
\normalsize{$^\ast$To whom correspondence should be addressed; E-mail:  subella@mit.edu.}
}

\date{}

\newcommand{\quadpos}{\vp}
\newcommand{\quadrot}{\MR}
\newcommand{\quadrvel}{\boldsymbol{\Omega}}

\newcommand{\rotcolz}{\vb_z}

\newcommand{\optional}[1]{}%

\newcommand{\disturbance}{\vtheta}
\newcommand{\estDisturbance}{\bar{\disturbance}}

\newcommand{\twoliter}{two-liter\xspace}

\newcommand{\medkit}{med-kit\xspace}

\newcommand{\realsenseT}{RealSense T265\xspace}
\newcommand{\realsenseD}{RealSense D455\xspace}

\newcommand{\topspeed}{2.0~m/s\xspace}
\newcommand{\nflights}{180\xspace}

\newcommand{\movie}{\href{https://www.youtube.com/watch?v=HF4M7TooqfE&feature=youtu.be}{Movie 1}}
\begin{document} 

\baselineskip24pt

\maketitle 

\begin{sciabstract}

\vspace{-6mm}
\begin{abstract}
Contrary to the stunning feats observed in birds of prey,
aerial manipulation and grasping with flying robots still lack versatility and agility.
Conventional approaches using rigid manipulators require precise positioning and are subject to large reaction forces at grasp, which limit performance at high speeds. The few reported examples of aggressive aerial grasping rely on motion capture systems, or fail to generalize across environments and grasp targets.
We describe the first example of a soft aerial manipulator equipped with a fully onboard perception pipeline, capable of robustly localizing and grasping visually and morphologically varied objects.
The proposed system features a novel passively closing tendon-actuated {soft gripper} that enables fast closure at grasp, while compensating for position errors, complying to the target-object morphology, and dampening reaction forces. 
The system includes an onboard perception pipeline that combines a neural-network-based semantic keypoint detector with a state-of-the-art robust 3D object pose estimator, whose estimate is further refined using a fixed-lag smoother.
The resulting pose estimate is passed to a minimum-snap trajectory planner, tracked by an adaptive controller that fully compensates for the added mass of the grasped object. 
Finally, a finite-element-based controller determines optimal gripper configurations for grasping.
Rigorous experiments confirm that our approach enables dynamic, aggressive, and versatile grasping. We demonstrate fully onboard vision-based grasps of a variety of objects, in both indoor and outdoor environments, and up to speeds of \topspeed --- the fastest vision-based grasp reported in the literature. %
Finally, we take a major step in expanding the utility of our platform beyond stationary targets, by demonstrating motion-capture-based grasps of targets moving up to 0.3~m/s, with relative speeds up to 1.5~m/s.
 
\end{abstract}

\end{sciabstract}

\section{Introduction}
\label{sec:intro}

Current quadrotor platforms can fly at high speeds, maneuver with great agility, and carry a wide range of payloads, but their ability to \textit{interact} with the world remains limited.
Most quadrotor platforms sense their environment for applications such as navigation, inspection, or videography~\cite{idrissi2022quadreview,joubert2016drone,hassanalian2017dronesurvey,gupte2012quadsurvey}. 
However, many tasks require interacting with the environment, and quadrotors built to accomplish such tasks must handle complicated time-varying contact dynamics and have the ability to perceive task-relevant features of the environment. 
Recent work has
increasingly focused in this direction, including a number of perching
mechanisms \cite{nguyen2023soft, zufferey2022ornithopters, roderick2021bird}, and full
systems for applications such as contact-based
inspection~\cite{Bodie19arxiv}, and tree eDNA
collection~\cite{aucone2023scirob}. These tasks require contact with the
environment, but not necessarily changing the environment. Other work has explored using quadrotors for building structures~\cite{augugliaro2013building,augugliaro2014flight}, but requires external localization to function. Our work enables
quadrotors to take a more active role in shaping the world around them by
picking up and moving objects without the need for external motion capture infrastructure, expanding the kinds of tasks that can be
accomplished by such platforms. 

Building an aerial grasping system requires solving problems at the
intersection of mechanical design, perception, motion planning, control, and manipulation. 
There is extensive prior work toward solving each of these problems{} individually. Many
works focus on gripping or grasping mechanisms for attaching to conventional
drones \cite{nguyen2023soft,zufferey2022ornithopters,roderick2021bird,
Xu2023aerial,appius2022RAPTOR} or helicopters \cite{Pounds11icra}. Recent work
has examined a closer coupling between the drone and gripper design
\cite{ryll2022smors,nguyen2023soft}. 
The control of aerial manipulators has been studied in \cite{thomas2013avian,
spica2012aerial,Rossi17ral,Kannan2016control}, and recently for related tasks such as catching balls in
flight \cite{yu2023catch} and grabbing other drones out of the air
\cite{chen2022aerial}. 
Vision-based object pose estimation and tracking are widely studied problems in
robotics, and many existing methods are relevant for estimating the target's
position for aerial manipulation; notable examples applied to aerial grasping
include \cite{sun2022ICK}.
While each of these works reports promising advances in individual subsystems, \emph{aggressive and versatile} aerial manipulation still remains out of reach, and prior work falls short of presenting a 
system that can perform aggressive grasping without strong assumptions about the target object and the environment.

Combining perception, control, and manipulation to achieve {aggressive and versatile} aerial manipulation 
presents several challenges.
First, conventional approaches using rigid manipulators require precise positioning and are subject to large reaction forces at grasp, which limit performance at high speeds. 
Second, in
contrast to systems that passively sense the environment, a quadrotor
manipulating objects in the environment must adapt to changing dynamics, in particular since the 
added mass of the grasped object might be non-negligible compared to the mass of the drone itself. 
Third, if
such systems are to be generally useful, they must also function without
perfect state information from external motion capture systems; therefore they must rely on onboard perception and their perception system has to be resilient to the manipulator partially occluding the onboard camera.
 These problems must be 
 solved in real-time, to enable efficient operation, and with the limited onboard computation, 
 to circumvent the need for an external infrastructure.

The few reported examples of aggressive aerial grasping rely on motion capture systems, or fail to generalize across environments or grasp targets. For instance, Thomas et al.~\cite{thomas2013avian} demonstrate aggressive grasping in a motion capture system, but they assume a suspended target to avoid the risk of undesired contact forces with the surface the object is lying on. Fishman et al.~\cite{Fishman21iros-softDrone2} demonstrate dynamic grasping with a soft aerial gripper, but the system requires external localization for the quadrotor and target.
Existing
vision-based aerial manipulation systems reduce errors by using a motion-capture system for the drone's self-localization~\cite{bauer2022autonomous}, by flying at slower
speed~\cite{lin2019autonomous,ramonsoria2020grasp}, or by making strong assumptions about the target 
shape that cannot be extended to arbitrary
objects~\cite{seo2017aerial,thomas2014grasping_perching}. 
None of these works
combines vision-based drone localization and target pose estimation, the ability to
pick up targets with meaningful forward velocity, and fully onboard computation.

\begin{figure*}[tbp]
    \centering
    \includegraphics[width=\textwidth]{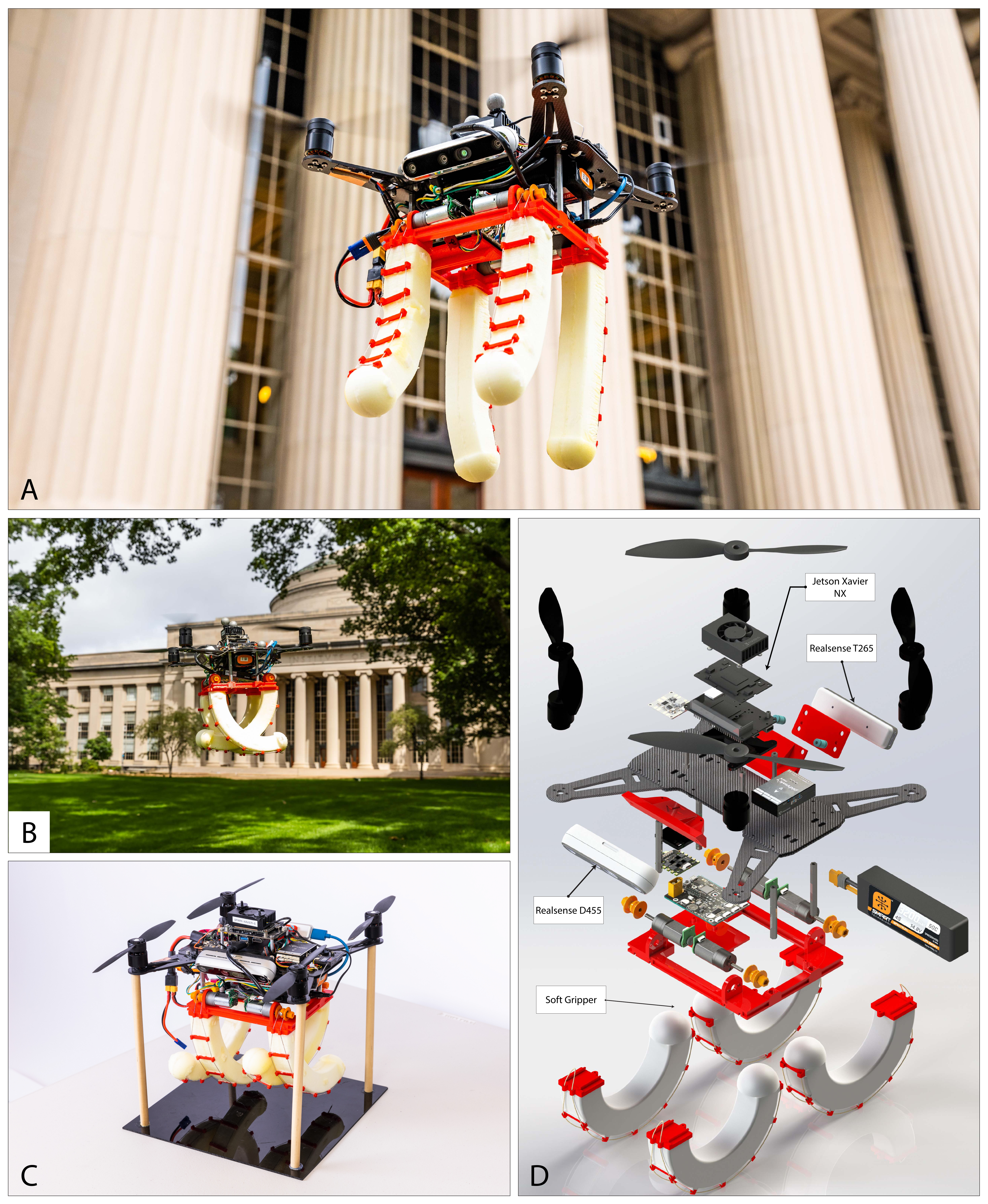}
    \caption{\textbf{Overview of the Soft Drone.}
    (\textbf{A-B})~The proposed Soft Drone platform flying outdoors.
    (\textbf{C})~Front view of the Soft Drone platform, with the soft gripper in the passively closed state.
    (\textbf{D})~Exploded CAD view of the quadrotor and gripper, including onboard sensing and computation.
    \label{fig:overview}
    }
\end{figure*}

{\bf Contribution.} We posit that aerial grasping can be improved by focusing on
\emph{soft} grasping mechanisms rather than trying to achieve extremely precise
positioning with a rigid manipulator. We draw inspiration from biological systems (\eg birds), which  utilize a combination of rigid and soft tissues to achieve unparalleled agility and robustness~\cite{Bern19rss}. Soft robotic
grippers can passively conform to the grasped object, reduce the need for
explicit grasp analysis, and weaken the coupling of flight and manipulation
dynamics; this is an example of \emph{morphological computation}, \ie the
exploitation of passive mechanical elements to supplement explicit
control~\cite{Rus15nature}. In this work, we combine a rigid quadrotor platform with a soft robotic gripper. Rigid and soft robots can operate together synergistically~\cite{stokes2014hybrid} as hybrid systems that get the best of worlds:
our system combines the speed and agility of a rigid quadcopter with the natural compliance and robustness of a soft robotic gripper, and harnesses control methodologies that enable its rigid and soft components to work together.

 Our first contribution is to develop a \textit{soft, quadrotor-mounted
gripper} with passively closing and tendon-actuated foam fingers. 
Our design enables fast closure at grasp, while compensating for positions errors, complying to the target-object morphology, and dampening reaction forces. 
Our second contribution is to develop a real-time and fully onboard perception pipeline for drone and target state estimation.
Our perception pipeline combines a neural-network-based semantic keypoint detector with a state-of-the-art method for robust 3D object pose estimation and a fixed-lag smoother. 
Finally, we integrate a minimum-snap grasp trajectory planner with an adaptive controller for the drone and a finite-element-based approach for the soft gripper control. 
The adaptive controller is able to adjust to 
quickly changing dynamics, caused by the grasped object and related aerodynamic effects (e.g., ground effect and down-wash resulting from the grasped object); the finite-element-based approach computes optimal configurations for the soft gripper that are key for high-speed grasping.

We evaluate our soft drone with onboard perception when grasping a variety of objects in both indoor and outdoor environments.
We present extensive experimental results across \nflights flight tests.
In contrast to prior work, our system performs all computation on board, can fly with or without a motion capture system, and can perform aggressive grasping of static targets at up to \topspeed.
Moreover, we show that when operating within a motion capture system the soft drone can also grasp objects from \emph{moving} platforms: a quadruped robot carrying a \medkit moving forward at 0.3 m/s, and a rotating turntable with a relative grasp speed of 1.5 m/s.

\section{Results}
\label{sec:results}

We have designed and built a combined quadrotor and soft gripper platform (\cref{fig:overview}A,B) able to aggressively grasp different object types at high speeds.
The gripper mechanism consists of four flexible, tendon-actuated foam fingers to provide robust grasp performance.
The quadrotor uses two onboard cameras (\cref{fig:overview}C,D) to estimate its own state and a target object's pose, which enables operation without external localization infrastructure such as a motion-capture system.
All perception, planning, and control algorithms run onboard the quadrotor.
To demonstrate the effectiveness of our system, we performed \nflights flights, where we measured grasping performance and provided an ablation of our main design choices.
In this section, we provide an overview of our system (Section~\ref{sec:systemOverview}) and then discuss the results of our flights with both 
static (Section~\ref{sec:staticGraspPerformance}) and moving targets (Section~\ref{sec:dynamicTarget}).  
More technical details are postponed to the Materials and Methods section, and visualizations of the system and experiments are shown in \movie.

\subsection{System Overview}
\label{sec:systemOverview}

 \begin{figure*}[tbp]
    \centering
    \includegraphics[width=\textwidth]{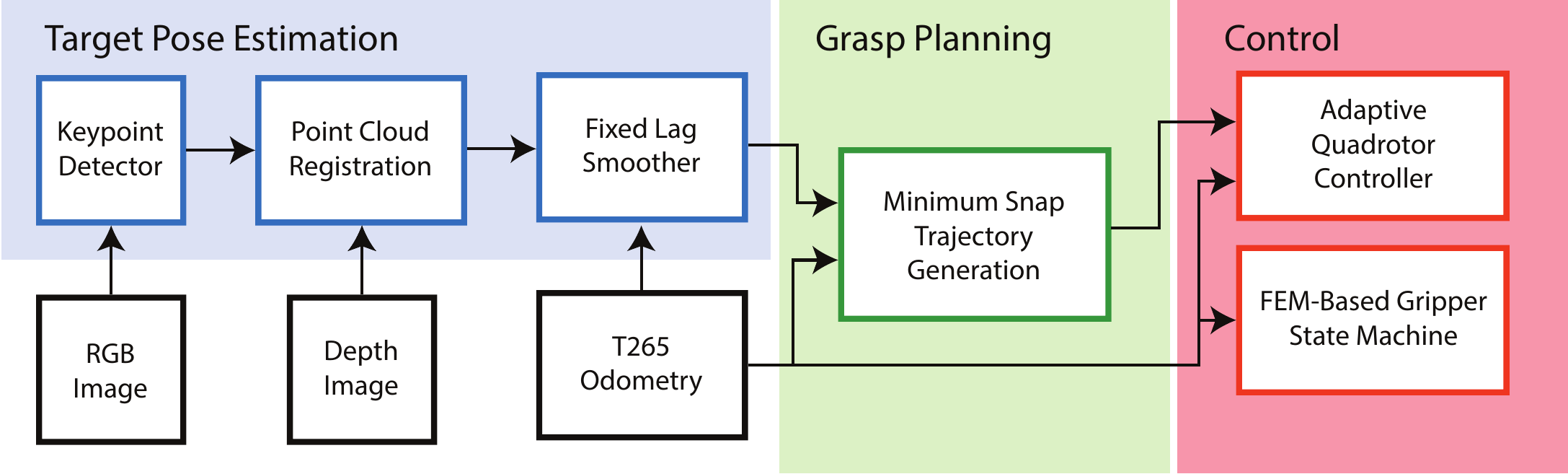}
     \caption{\textbf{Grasping pipeline overview.} An accurate estimate of the target and drone's pose is first identified and used to plan the polynomial grasp trajectory. This trajectory is tracked with an adaptive quadrotor controller, and the soft gripper states are dictated by an FEM-based optimization.} \label{fig:flow_chart}
\end{figure*}

A successful object grasp requires the drone to detect and estimate the object's pose, plan a feasible grasping trajectory, and track that trajectory while facing complicated contact dynamics and other external disturbances.

Our system (\cref{fig:flow_chart}) is designed to perform fully onboard perception, using a \realsenseD color and depth camera for object pose estimation and a \realsenseT stereo fisheye camera for visual-inertial odometry (VIO) for drone state estimation~\cite{huang2019vio}.
After estimating the target's and drone's pose in the global frame,\footnote{While some aerial grasping approaches based on visual servoing avoid the need for an explicit world-frame target estimate \cite{seo2017aerial,thomas2014grasping_perching}, we form a global estimate of the target's pose to remove the constraint that the target must always be visible in the camera's field of view, which is a restricting assumption in particular right before grasping.} the quadrotor plans a minimum-snap polynomial trajectory~\cite{Mellinger11icra} connecting its initial hover point, a grasp point directly above the target, and a terminal hover point.
This reference trajectory is then tracked by an adaptive flight control law~\cite{Goodarzi15adaptive} which learns to compensate for the additional mass after grasp and for unmodeled aerodynamic effects.
The gripper's finger positions are controlled along the trajectory using a finite-element-based approach~\cite{Bern17iros,Bern19rss} governed by an objective functions that balances 
target object visibility and grasp robustness.
The object pose estimation and trajectory planning algorithms run on a Jetson Xavier NX, the low-level flight control runs on a Pixhawk micro-controller, and the visual-inertial odometry pipeline runs on the \realsenseT camera module.

Below, we provide an overview of our soft gripper design (Section~\ref{sec:softGripperDesign}) and 
the onboard perception system (Section~\ref{sec:perceptionSystemDesign}), 
while we postpone the details to Materials and Methods.

\subsubsection{Soft Gripper Design and Modeling}
\label{sec:softGripperDesign}

\begin{figure*}[tbp]
    \centering
    \includegraphics[width=\textwidth]{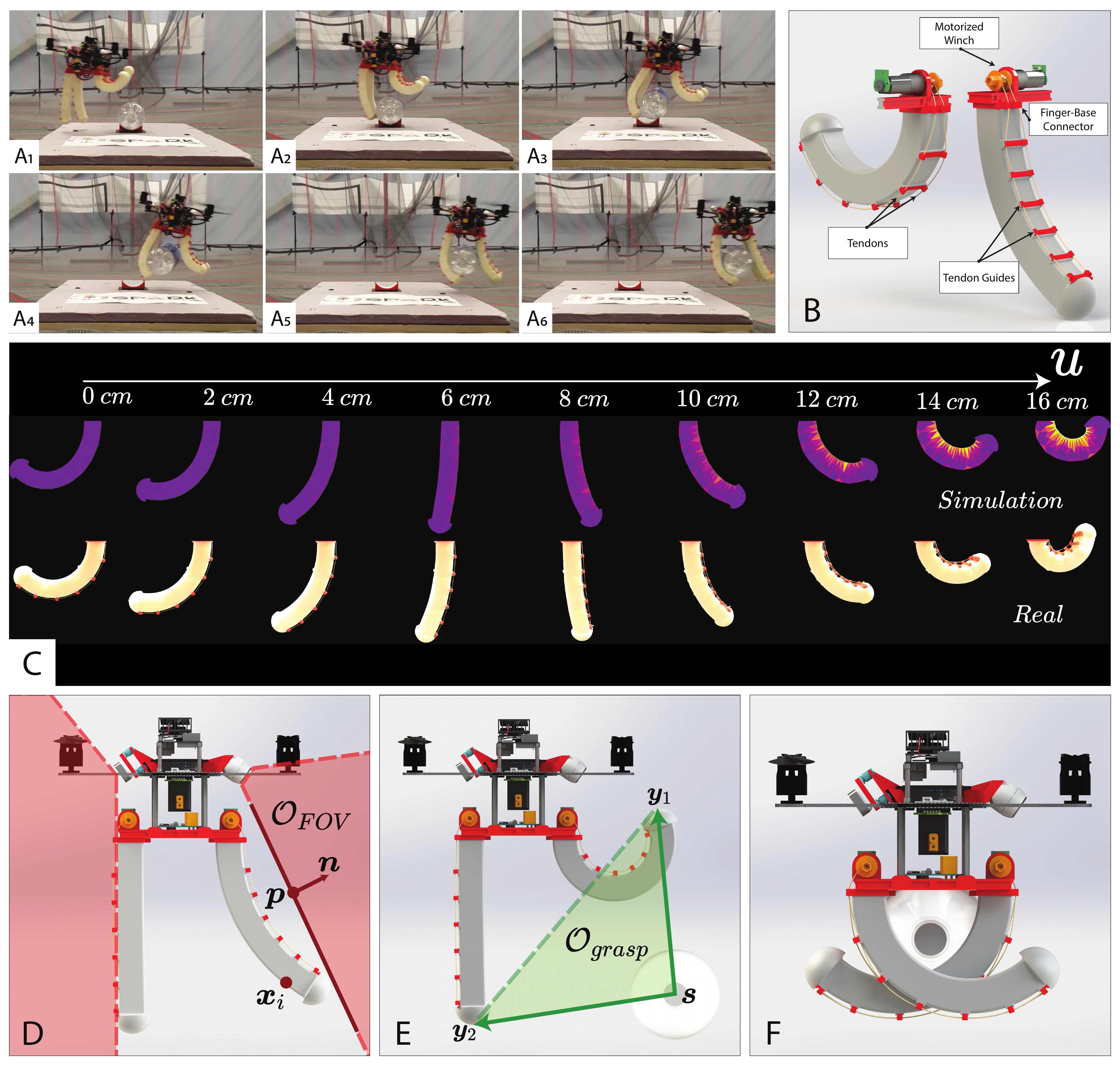}
    \caption{
    \textbf{FEM modeling and objective functions.}
    (\textbf{A})~Still frames from a grasp with over 2 m/s forward velocity. The compliance of the soft fingers mitigates the high reaction forces faced at this speed (\textbf{A\textsubscript{3}}) and the gripper configuration reduces the sensitivity of the gripper closing timing.
    (\textbf{B})~Soft finger in its passively closed state (left) and partially open state (right). The finger opening is actuated by a motorized winch contracting a tendon fixed to the tip of the finger and routed through guides.
    (\textbf{C})~The finger shape predicted by the FEM model based on cable retraction length $u$ (top) matches the actual finger shape for the same cable length (bottom).
    (\textbf{D})~Target observation state: the field-of-view objective, $\mathcal{O}_{\text{FOV}}$, drives the mesh nodes, including the node with position $\mathbf{x}_i$, out of the half space bounded by the plane defined by point $\bm{p}$ and normal $\bm{n}$ and corresponding to the cameras' frustum.
    (\textbf{E})~Pre-grasp state: the grasp margin objective, $\mathcal{O}_{\text{grasp}}$, maximizes the area enclosed by the grippers fingertips $\mathbf{y}_1$ and $\mathbf{y}_2$ and the target point $\mathbf{s}$.
    (\textbf{F}) Gripper closed configuration: the fingers return to their passively closed state.
    }
    \label{fig:fem}
\end{figure*}

Eliminating all estimation and tracking errors from the perception and control systems is infeasible, so the grasp mechanism must be robust to imprecise positioning to enable reliable grasps.
Collisions between the fingers and the ground present a particular challenge.
Small amounts of error in the planned or executed trajectory lead to the manipulator contacting the ground.
With a traditional rigid manipulator, these unexpected contact forces induce a moment on the quadrotor that results in it rapidly pitching forward, leading to a crash or failed grasp~\cite{Fishman21iros-softDrone2}.

We propose to solve this issue with \textit{compliant} fingers for picking up the object (\cref{fig:fem}A).
Such fingers weaken the coupling between ground contact and drone dynamics, enabling steadier flight in the event of ground contact (\movie).
Soft fingers also provide robustness for grasping the object: they conform to the object morphology and do not require precisely chosen grasp points.

Our finger design is made of foam molded in a closed position.
Soft foams, especially castable expanding polyurethane, have been recently harnessed to produce highly-deformable soft robots with favorable strain-stress ratios \cite{kastor2020design}.
A cable connected via eyelets along the outside edge of the finger controls its shape (\cref{fig:fem}B).
As the motor at the base of the finger turns, the cable shortens and pulls the finger into an open position.
This passively closed finger design has the advantage that the closing speed is limited only by the stiffness of the finger and no-load speed of the motor, in addition to lower power draw when the fingers are closed.
The resulting finger design is mechanically quite simple compared to more traditional rigid finger designs that require actuation at each joint.
In comparison to our previous soft finger design~\cite{Fishman21iros-softDrone2} that required two motors working in synchronization to open and close each finger, our current design has half the number of motors and is much easier to fabricate and control.
Now each finger only has a single degree of freedom, yet it can assume nontrivial shapes (\cref{fig:fem}C). 

The positioning of the fingers is important for a successful grasp.
We model the body of the gripper as a finite element mesh, and the cables as unilateral springs running through via points in the mesh. Given a feasible choice of real-world control inputs, e.g., motor angles $\bm{u}$, this approach to soft robotic modeling can accurately predict the real finger's deformed shape. This is done by minimizing the total energy of the system, to find a statically stable deformed mesh position, following the approach in~\cite{Bern19rss,Bern17iros}:
\begin{equation}
\bm{x}(\bm{u}) = \arg\min_{\bm{x}} E(\bm{u}, \bm{x}).
\end{equation}
In addition to helping us predict the gripper's motion, the FEM-based model can be harnessed in a nested optimization to do control.
In this work, we find optimal control inputs by minimizing a suitable objective function:
\begin{equation}
\bm{u}^* = \arg\min_{\bm{u}} \bigg(\mathcal{O}_{\text{grasp}}(\bm{x}(\bm{u})) + \mathcal{O}_{\text{FOV}}(\bm{x}(\bm{u}))\bigg),
\label{eq:objective}
\end{equation}
where the objective is designed to produce control inputs that deform the gripper 
into a shape $\bm{x}(\bm{u}^*)$ that maximizes the area enclosed by the fingertips and the target's centroid (as quantified by the term $\mathcal{O}_{\text{grasp}}(\bm{x}(\bm{u}))$ in Eq.~\ref{eq:objective}, see Materials and Methods
 for a complete mathematical expression) while also not occluding the quadcopter's field of view (as quantified by the term $\mathcal{O}_{\text{FOV}}(\bm{x}(\bm{u}))$). 
 The exact choice of objective depends on the different phases of the grasp, as discussed below.
 The optimization is performed offline to obtain optimal motor angles, which are then commanded during the grasp procedure.

When the drone is planning the grasp trajectory (\cref{fig:static_timelapses}A-1), the front fingers should not be in the front camera's field of view or they may obscure the object (\cref{fig:fem}D).
The back fingers also need to stay out of the navigation camera's field of view, to avoid interfering with the drone localization system.
 We refer to this configuration as \textit{target observation state} (\cref{fig:static_timelapses}B-1), and set $\mathcal{O}_{\text{FOV}}(\bm{x}(\bm{u})) =  \mathcal{O}_{\text{FOV}_R}(\bm{x}(\bm{u})) + \mathcal{O}_{\text{FOV}_F}(\bm{x}(\bm{u}))$ as the sum of the FOV constraints for the rear (T265) and front (D455) cameras.

Right before grasping the target (what we call the \textit{pre-grasp state}, \cref{fig:static_timelapses}B-2), we can remove the front camera FOV constraint because we have global knowledge of the target's location. Therefore, $\mathcal{O}_{\text{FOV}}(\bm{x}(\bm{u})) =  \mathcal{O}_{\text{FOV}_R}(\bm{x}(\bm{u}))$, and the front fingers can reach the unconstrained grasp configuration, which increases the grasp volume and greatly improves our position error margins (\cref{fig:fem}E).
We note that this objective function also results in the rear fingers pointing downward; if the fingers are closed too late, the target will still be caught up by the back fingers and a successful grasp may still be possible (\cref{fig:fem}A). 

To initiate the grasp closure, the tendons are simply commanded back to their rest positions and the fingers return to the passively closed state, thus enclosing the target (\cref{fig:fem}F, \cref{fig:static_timelapses}B-3).

\subsubsection{Perception System Design}
\label{sec:perceptionSystemDesign}

\begin{figure*}[tbp]
    \centering
    \includegraphics[width=\textwidth]{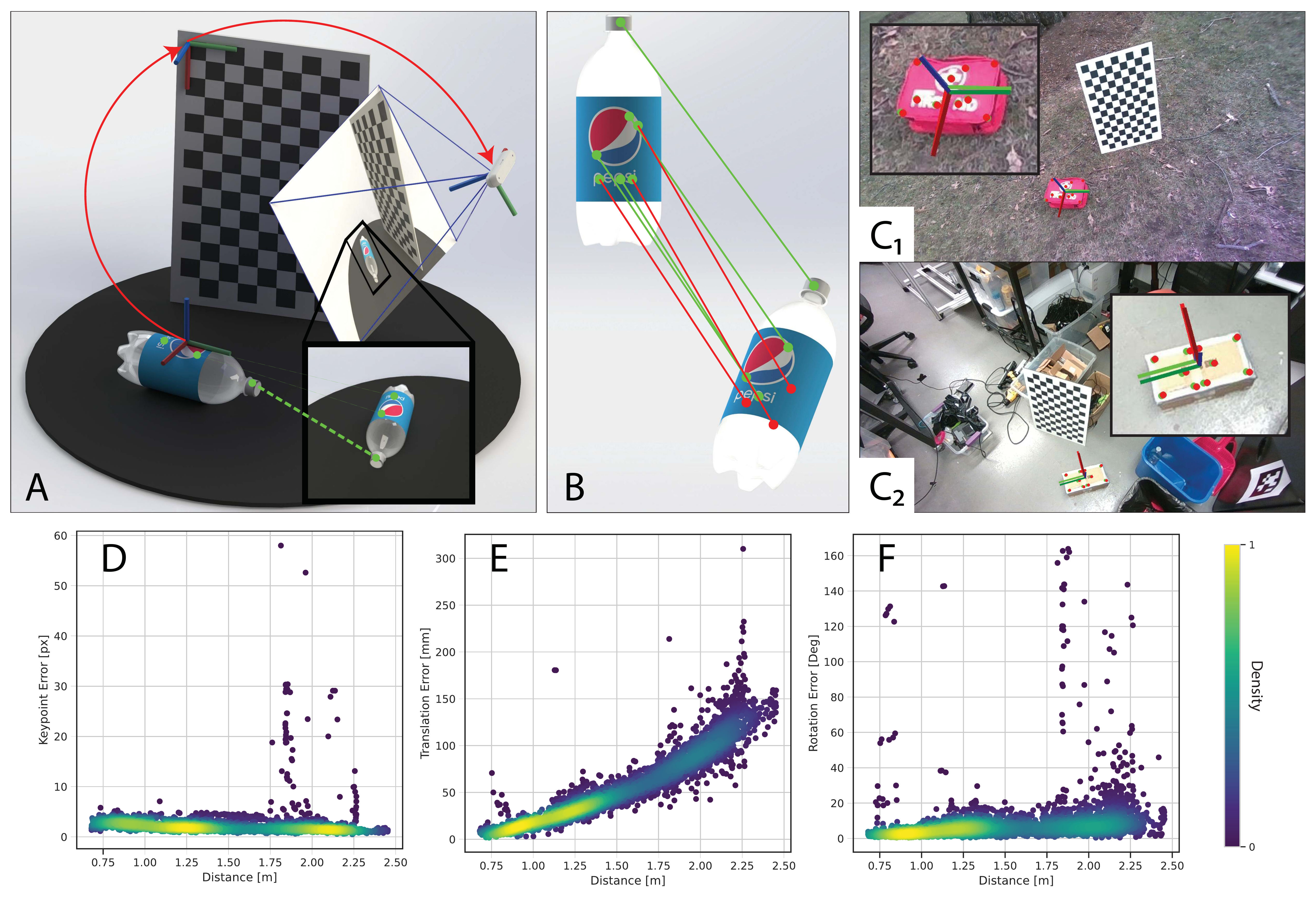}
    \caption{
    \textbf{Keypoint-based object pose estimation.}
    (\textbf{A})~Keypoints are labeled to generate training data for the semantic keypoint detector, 
    for each object of interest. A checkerboard in the background of the collection area enables calculation of camera pose, which accelerates and partially automates the labeling process. A Resnet-18 neural network~\cite{trtpose} is trained to predict object keypoints based on this dataset.
    (\textbf{B})~The 3D keypoints predicted by the perception system are matched against the corresponding keypoint locations on the object CAD model. A rigid transformation computed between the observed and CAD keypoints gives an estimate of the relative pose between the object and the camera.
    (\textbf{C})~The keypoint and pose detection system generalizes across environments and targets. The red points are predicted keypoints, the green points are ground-truth keypoints, the dark colored axes indicate the estimated pose, and the light colored axes denote the ground-truth pose.
    (\textbf{D-F})~Error characteristics for the  proposed perception system, showing pixel-space, translation, and rotation error depending on distance between the target object and the camera for a \medkit object.
    }\label{fig:pose_plots}
\end{figure*}
Our perception system enables global position and orientation estimation of the target object (\cref{fig:pose_plots}).
It is important to estimate the target's orientation in addition to position, as some objects are much more difficult to pick up along certain axes (\eg the \twoliter bottle in \cref{fig:pose_plots}B). 
The perception pipeline assumes that the shape (more precisely, the CAD model) of the objects to be picked up is known ahead of time, but the framework is generalizable to arbitrary objects.

A deep-neural-network-based keypoint detector can be rapidly trained with an automated data collection tool (\cref{fig:pose_plots}A) and predicts a set of keypoints in each color image collected by the onboard Realsense 455 camera (\cref{fig:pose_plots}C); each semantic keypoint corresponds to a specific 3D point on the object CAD model.
The detected 2D points are mapped to 3D keypoints based on the image's depth channel.
The object's pose can be then estimated by aligning the detected keypoints to the corresponding points in the known object CAD model (\cref{fig:pose_plots}B). 
As there may be outliers in the semantic keypoint detection process, we employ a robust registration algorithm, namely TEASER++~\cite{Yang20tro-teaser}, to find the target pose.
Combining the camera-relative object pose with the quadrotor's own odometry estimate gives a global pose estimate for the target object (\cref{fig:pose_plots}C). 
In our implementation, we use the visual-inertial odometry computed by the T265 RealSense camera as the quadrotor's state estimate. 

\cref{fig:pose_plots}D-F show the pixel-space, translation, and rotation errors for the pose estimates produced by out perception pipeline for a \medkit on a test dataset held out from training. As the distance to the target increases, the translation estimate degrades while the keypoint and rotation errors remain roughly constant, indicating that our range is mostly limited by the depth camera's accuracy. 
In our flight tests, we manage this error by estimating the target's pose when the drone is within 1.25m of the target.
Additionally, these raw global pose estimates are filtered using a fixed-lag smoother to further improve accuracy; see
Materials and Methods for details.

Our system is capable of estimating the full 3D rotation of the target, which is necessary to plan grasps for targets on inclines or for determining the feasibility of a grasp. However, for most grasping applications, the target can be assumed to rest on a flat horizontal plane. Thus, only the target's yaw 
is used to plan trajectories.

\subsection{Grasping Static Targets}
\label{sec:staticGraspPerformance}

\begin{figure*}[tbp]
    \centering
    \includegraphics[width=\textwidth]{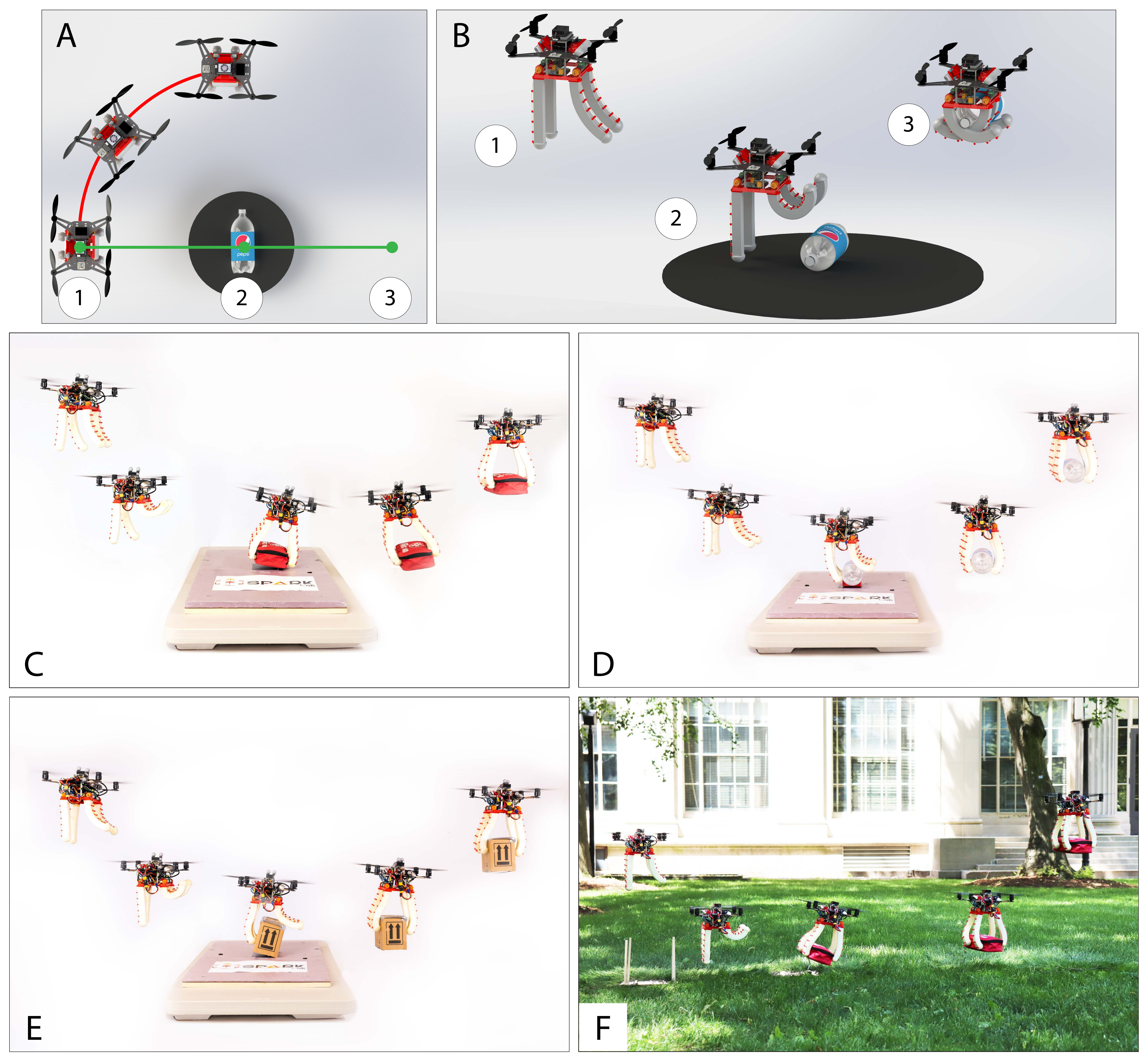}
    \caption{\textbf{Overview and timelapses of experiments with static targets.} (\textbf{A})~Experimental setup for the vision-based experiments. The red curve depicts the distribution of random start points, the green line the grasp trajectory; the numbers denote the grasp planning point $\textcircled{1}$, the grasp point $\textcircled{2}$, and the terminal point $\textcircled{3}$. (\textbf{B})~Configurations of the gripper states through the course of the grasp trajectory: $\textcircled{1}$ target observation state, $\textcircled{2}$ pre-grasp state, and $\textcircled{3}$ post-grasp state.
 (\textbf{C}-\textbf{E})~Static vision-based grasps at 0.5 m/s for \medkit, \twoliter bottle, and cardboard box, respectively. (\textbf{F})~Outdoor vision-based grasped of the \medkit at 0.5 m/s.}
    \label{fig:static_timelapses}
\end{figure*}

We conducted grasps of three distinct stationary objects (a \medkit, a \twoliter bottle, and a cardboard box, \cref{fig:static_timelapses}C-E) to demonstrate the versatility of the proposed soft gripper and vision system and quantify the effects of various sources of error.
In our tests, upon takeoff, the drone flies to a randomized starting point from which it can see the target, as shown in \cref{fig:static_timelapses}A.
Once the perception system has an estimate of the target's pose, it flies to a point such that the forward direction of the drone is aligned with a predefined grasp direction for the target. (\cref{fig:static_timelapses}A-1,B-1).
From that starting point, the quadrotor flies a trajectory through the grasp point (\cref{fig:static_timelapses}A-2,B-2) that terminates at an endpoint away from the target (\cref{fig:static_timelapses}A-3,B-3).
The drone stops updating its estimate of the target pose once it has started the grasp trajectory, relying on its initial estimate of target pose and updated estimate of its own state in the global frame in order to fly to the grasp point.

\subsubsection{Grasp Success Rate}
\begin{figure*}[tbp]
    \centering
    \includegraphics[width=\textwidth]{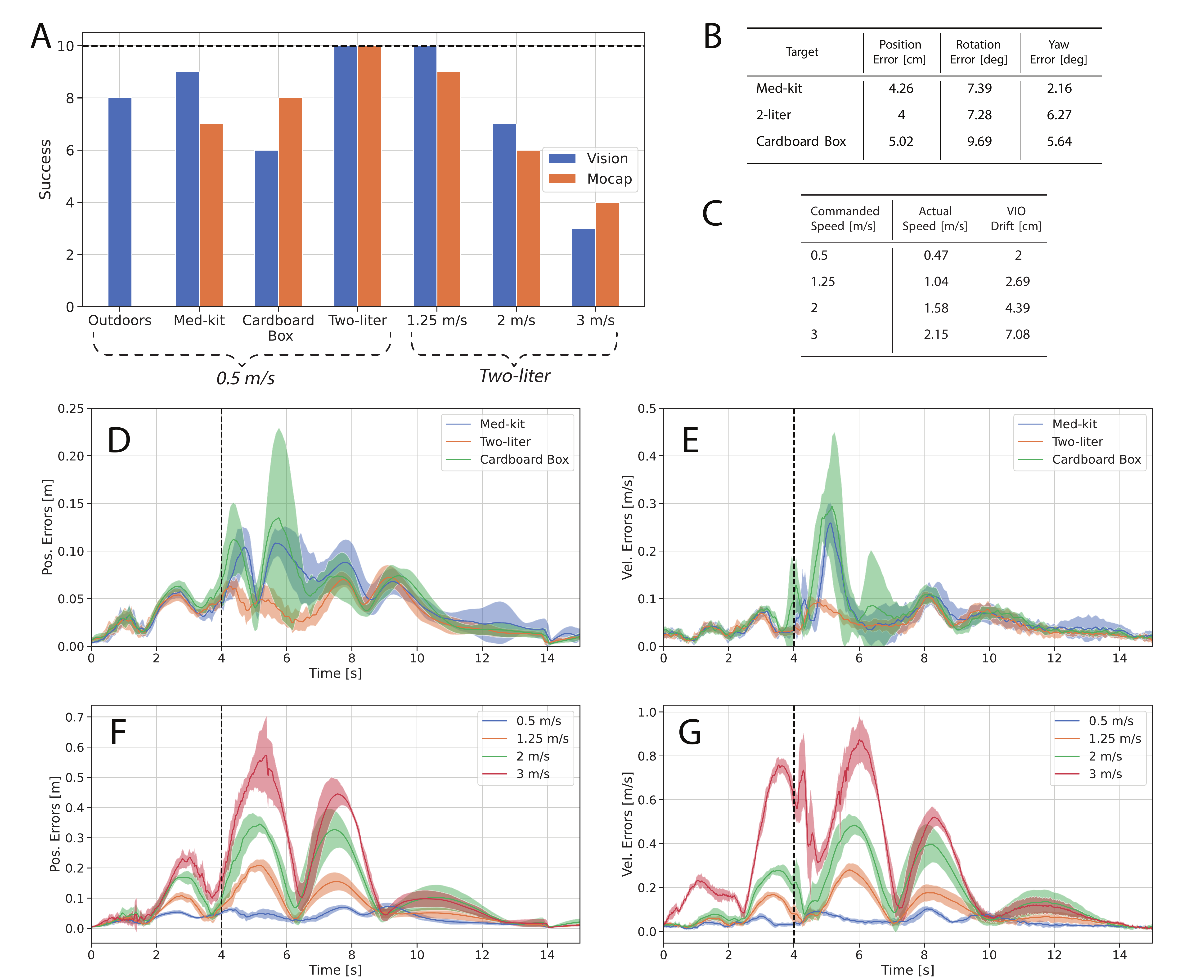}
    \caption{
        \textbf{Static grasp experiments results.}
        (\textbf{A})~Success rate for each experiment, comparing motion-capture-based performance (when available) with onboard-vision-based performance. 10 flights were conducted for each experiment, as indicated by the horizontal dashed line.
        (\textbf{B})~Filtered target pose estimation errors for each object at the time of trajectory planning, averaged across all runs.
        (\textbf{C})~Actual forward velocity as measured by motion capture and VIO drift for varying commanded speeds, averaged across runs.
        (\textbf{D-E})~Position and velocity tracking errors over the grasp trajectory for the three objects.
        (\textbf{F-G})~Position and velocity errors for four different speeds. The vertical dashed line denotes the time of grasp and the shaded regions represent one standard deviation from the mean. The errors here are computed as the mismatch between the vision-based estimate and desired setpoint.}\label{fig:vision_static}
\end{figure*}

In this first set of tests, the reference trajectories were chosen to have a forward velocity of 0.5~m/s at the grasp point.
A grasp is considered a success if the target remains grasped until the drone lands.
For all tests, the target is secured to a horizontal surface with weak magnets to prevent the downdraft of the quadrotor displacing the target pre-grasp.
Under these conditions, the system achieves a success rate of 9/10, 6/10, and 10/10 for the \medkit, cardboard box, and \twoliter bottle, respectively, and with fully onboard vision ({\cref{fig:vision_static}A}). 
We repeated these experiments with a motion capture system to determine the extent that our perception system affects grasp success.
The motion capture baseline performs similarly to the vision-based approach (\cref{fig:vision_static}A), slightly under-performing for the \medkit, improving for the cardboard box, and performing equally for the bottle.
Our results show that our vision system is able to consistently match the motion capture results at this speed, indicating that the gripper mechanism is robust against the additional error caused by the perception pipeline. 

The biggest contributors to grasp performance difference among objects are their mass and surface morphology. 
The \twoliter bottle's cylindrical geometry and low mass (60g) enable consistent enveloping grasps.
The uneven, roughly concave edges of the \medkit provide good grasp surfaces, but the high mass (148g) makes sustaining the grasp more difficult.
The cardboard box's tall, straight walls combined with its relatively high mass (115g) force the gripper to rely on large pinching force to secure the grasp, making it the most difficult target to grasp.

\subsubsection{Error Analysis}
We collected ground-truth pose data for the drone and target with a motion-capture system 
in order to understand the relative contribution of errors from the target estimation pipeline, the VIO estimate, and the trajectory tracking to the system's performance.

The filtered pose estimate for each target (\cref{fig:vision_static}B) at the start point of the trajectory has at most 5~cm translation error and less than 10 degrees of rotation error. Furthermore, only the target's yaw is needed for grasp planning and at most has roughly 6~degrees of error.

The average error between the quadrotor's actual position and VIO-based estimate (\cref{fig:vision_static}C) up until the grasp point is about 2~cm, for a forward speed of 0.5~m/s.

Finally, the trajectory tracking performance (\cref{fig:vision_static}D,E) before the grasp point is comparable for all three objects, with approximately 5~cm position error, and 0.05~m/s velocity error.
One cause of tracking error before the grasp time is the presence of the \emph{ground effect}, where the drone experiences greater thrust efficiency when approaching a horizontal surface~\cite{Bernard18aagnc-groundEffect}, thus raising the drone vertically above its desired setpoint (Figure S1).
This is partially compensated by the adaptive controller, and the ability of the soft fingers to safely contact the ground reduces the risk from overcompensation.
Apart from specific aerodynamic effects, our platform has a low thrust-to-weight ratio, which limits its control authority and invokes slight tracking errors even in normal flight conditions. 

After the grasp, the higher mass of the \medkit and cardboard box lead to increased tracking errors primarily in the vertical direction (Figure S1), which reduce over time as the adaptive controller learns to compensate. 
It is important that both pre- and post-grasp errors are minimal for a successful grasp: high pre-grasp errors may cause the drone to be misaligned and miss the target, while high post-grasp errors could cause unintended collisions with the environment (\eg floor) and destabilize the drone.

Further breakdown of errors across separate axes is presented in Figure S1, Tables S1-3, and Supplementary Text S1, and information on how these error metrics are measured is provided in Section~\ref{sec:error_sources}.

\subsubsection{Grasp Speed Analysis}
\label{sec:speed_ablation}

We further evaluate the performance of our system by increasing the desired forward grasp velocity to 1.25~m/s, 2~m/s, and 3~m/s for the \twoliter bottle.
We conducted ten flights for each speed with the full vision-based system, and repeated another ten flights for each speed in motion capture as a baseline.
Our system is capable of grasping the bottle with full vision at over 2 m/s with a success rate of 3/10 (\cref{fig:vision_static}A).
To our knowledge, this is the fastest  fully vision-based aerial grasp reported in the literature. 

The increases in speed are accompanied by larger tracking errors in both position and velocity (\cref{fig:vision_static}F, G). 
Because there exists a large discrepancy between the desired and actual speeds at grasp, we report the drone's average actual grasp speed as measured by the motion capture system in (\cref{fig:vision_static}C).
This error can be attributed to lower control authority at higher speeds; our platform operates near its maximum payload capacity and struggles to accelerate quickly.
However, the error is mostly along the longitudinal direction (i.e., grasp or forward axis), with the lateral and vertical errors remaining low regardless of speed.
Longitudinal tracking errors have little effect on our grasp success as the gripper control is informed by the current position of the drone, not its desired position.
As long as the drone's state estimate remains accurate up until the time of grasp, the grasp will be triggered at the appropriate time. 
We validate this by analyzing the VIO drift for the varying speeds (\cref{fig:vision_static}C).
VIO drift does increase with grasp speed, remaining manageable for desired speeds less than 2 m/s, but reaching a substantial 7~cm for a desired speed of 3 m/s. As reflected in \cref{fig:vision_static}A, this increase in VIO error proves to have a non-negligible effect on grasp performance.
Tables S1-S3 further break down VIO error per axis for each speed.

\subsubsection{Outdoor Grasps}
A central motivation for our onboard perception system is the ability to operate in places where a localization infrastructure is unavailable.
To emphasize this capability, we demonstrate the quadrotor picking up the \medkit in an outdoor field (\cref{fig:static_timelapses}F) at 0.5 m/s. 
Grasping objects in outdoor environments presents several additional challenges, including unpredictable wind disturbances, varying illumination conditions, and visually diverse surroundings.
Furthermore, the sustained ground plane introduces an additional challenge compared to the indoor flights where the target was placed on a raised pedestal.
Post-grasp, the delay in adapting to the target's mass causes the drone to drag across the ground temporarily which increases the chance of failure. To combat this, we introduce a feedforward acceleration impulse that counteracts the immediate vertical disturbance post grasp and briefly raises the drone up from the ground plane. Afterwards, the drone continues tracking the nominal trajectory. We also note that the starting point of the drone was left fixed, rather than randomized as in the indoor experiments.
We demonstrate a success rate of 8/10, and, to our knowledge, the first instance of aggressive manipulation in an outdoor environment.

\begin{figure*}[tbp]
    \centering
    \includegraphics[width=\textwidth]{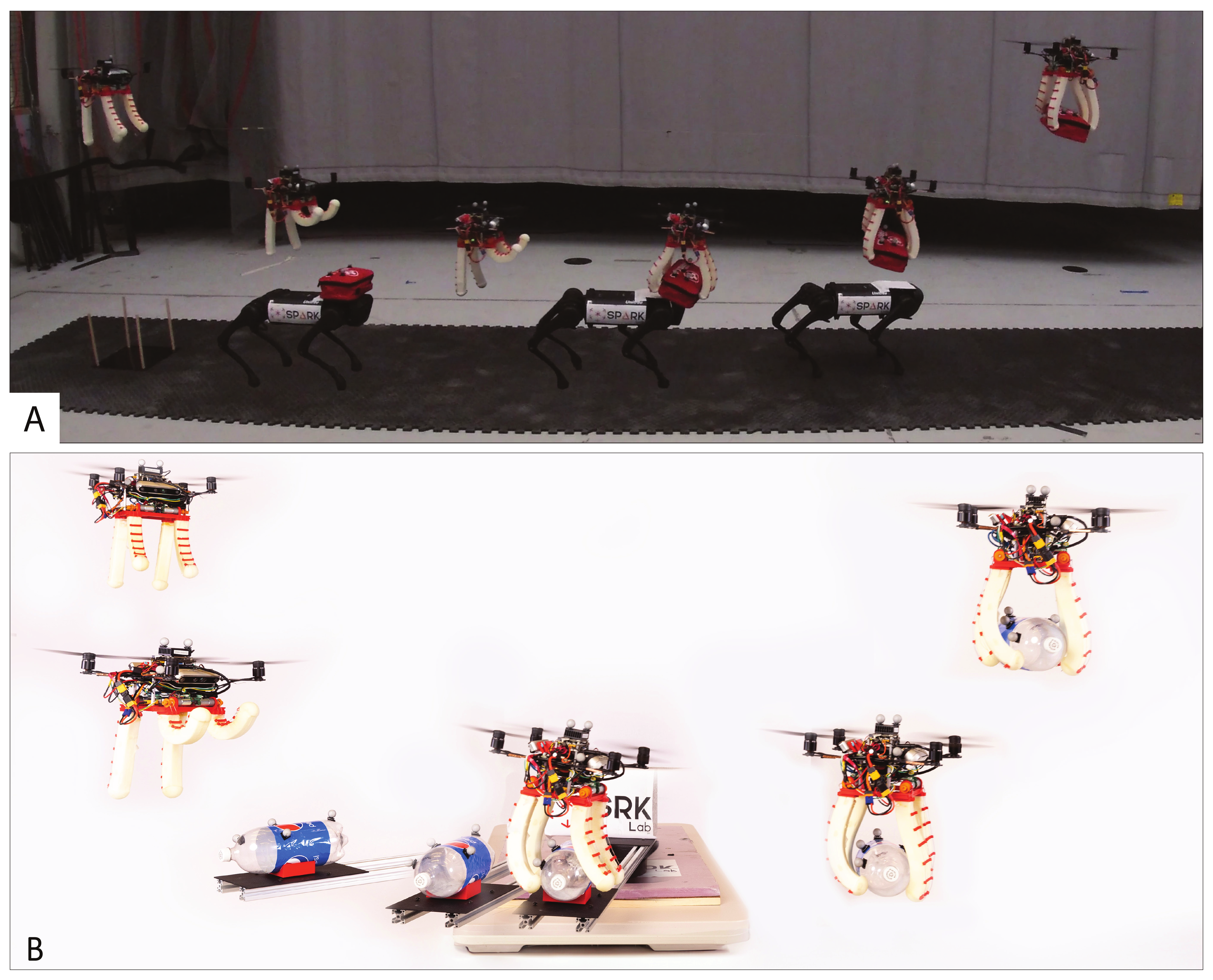}
    \caption{\textbf{Timelapses of experiments with moving targets.} (\textbf{A}) Moving grasps of \medkit attached to a quadruped robot that has forward velocity of 0.3~m/s and with a relative grasp speed of~0.1 m/s.
     (\textbf{B})~Moving grasps of \twoliter bottle on a turntable moving with a tangential speed of 0.08~m/s and with a relative grasp speed of 0.5~m/s; see \movie for more visualizations.}
    \label{fig:moving_timelapses}
\end{figure*}

 \begin{figure*}[tbp]
    \centering
    \includegraphics[width=\textwidth]{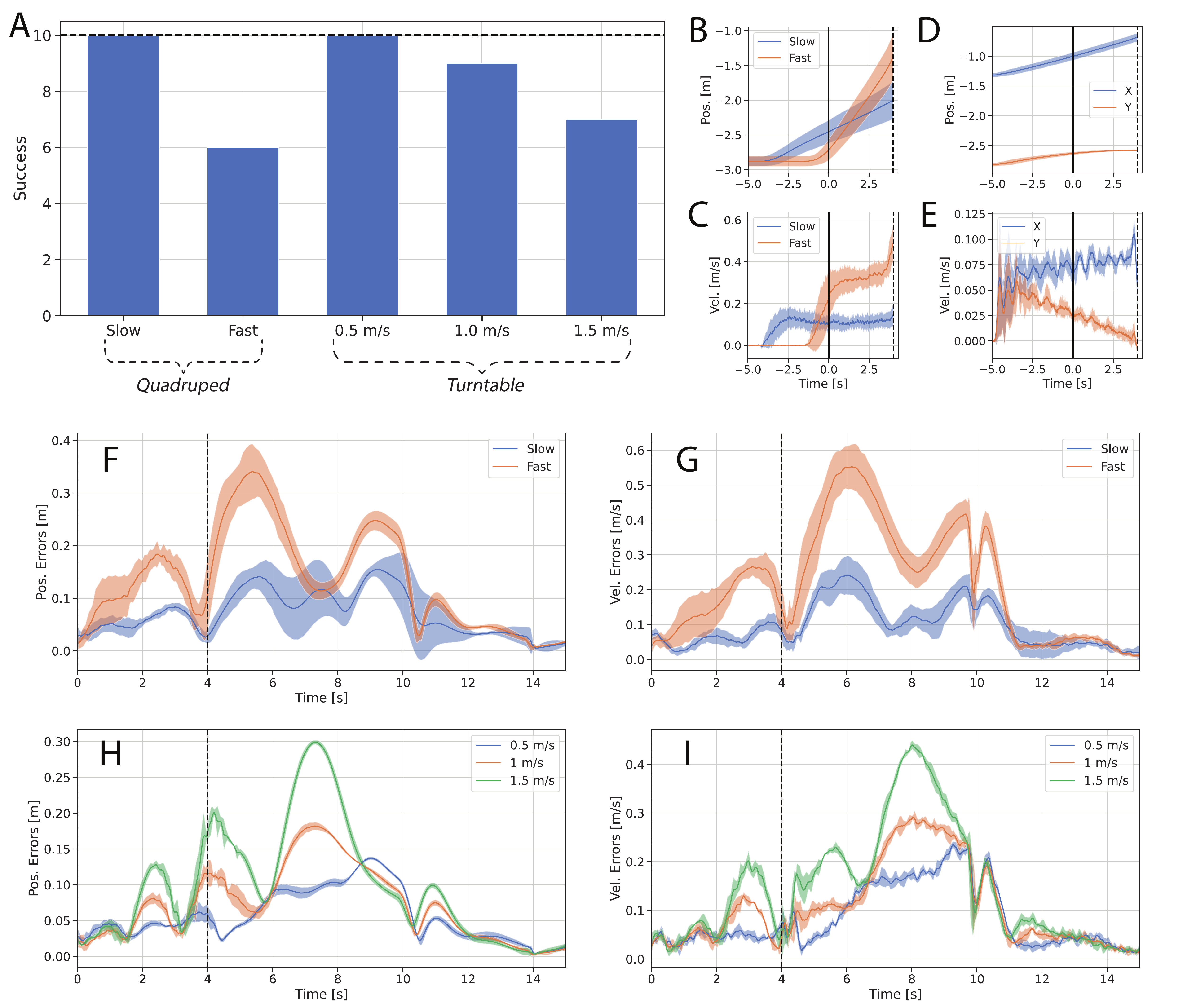}
    \caption{\textbf{Moving grasp experiments results.} 
    (\textbf{A})~Success rates for each experiment, including different moving platforms (Quadruped and Turntable) and increasing speeds.
    (\textbf{B-C})~Position and velocity of the target for the quadruped experiments.
    (\textbf{D-E})~Position and velocity of the target for the turntable experiments. 
    Here, $X, Y$ represent the axes in the inertial floor plane.
    (\textbf{F-G})~Position and velocity tracking errors for the quadruped experiments.
    (\textbf{H-I})~Position  and velocity tracking errors for the turntable experiments. For all plots, the solid black line denotes the start of the grasp trajectory, and the dashed black line is the time of grasp (after which the target pose is unavailable).}\label{fig:moving}
\end{figure*}

\subsection{Grasping Moving Targets}
\label{sec:dynamicTarget}

In this section we investigate
grasping \emph{moving} targets at high velocities, or with high relative velocity between the drone and the target.
These maneuvers are common in biological systems, where birds of prey must quickly capture their prey before it can escape.
In robotics applications, this capability can increase efficiency by removing the need to decelerate before grasping as well as minimize the time the drone must spend near the ground in dangerous environments. 

Grasping moving targets presents additional challenges on top of the ones observed with static targets.
When an object has rotational velocity, the desired grasp direction also rotates.
This requires more lateral tracking accuracy than is needed for the static cases.
Dynamic objects also have uncertainty in their motion model, which we observe when trying to grasp an object from a moving quadruped. 
While our system cannot simultaneously handle errors from perception and dynamics, we demonstrate the ability to aggressively grasp moving objects with the aid of motion capture for drone and target localization.

Our first set of experiments with moving targets consists of the \medkit mounted on top of a Unitree A1 quadruped robot (\cref{fig:moving_timelapses}A). 
A human operator steers the quadruped in a roughly linear motion (\cref{fig:moving}B).
Even though the trajectory is nominally linear and constant velocity, there is noticeable variation.
We conducted 10 flights at a slow quadruped forward velocity (roughly 0.15 m/s) and a fast forward velocity (roughly 0.3 m/s) (Fig.~\ref{fig:moving}C). 
The drone was commanded to grasp the med-kit with a relative forward velocity of 0.1 m/s. The drone achieved 10/10 success for the slow speed, and 6/10 for the fast (\cref{fig:moving}A). This difference can primarily be attributed to the increased tracking errors seen at grasp point for the higher speed (\cref{fig:moving}F,G), with noticeable increases in lateral and vertical errors (Figure S2, Tables S2, S3). 

For our second set of dynamic experiments, the \twoliter bottle sits on a lever arm attached to a motorized turntable (\cref{fig:moving_timelapses}B).
With a fixed rotation speed of the turntable, we perform 10 tests  for each desired relative grasp speeds of 0.5~m/s, 1.0~m/s, and 1.5~m/s.
For all experiments, the \twoliter bottle follows a circular arc (\cref{fig:moving}D) with a tangential speed of approximately 0.08~m/s (\cref{fig:moving}E).
Circular motion is challenging for our approach; because the drone tracks the target at a lever arm, the angular velocity of the bottle gets amplified into large translational velocities at the drone's setpoint.
Despite this, the drone exhibited high success rates for the three speeds (\cref{fig:moving}A). Similar to the the static speed analysis, the tracking errors increase with relative speed (\cref{fig:moving}H, I). However, these errors remain mostly concentrated in the longitudinal axis (Figure S2, Tables S2, S3), which has little effect on the grasp success. The rotational motion does induce higher lateral errors with speed (Table S2), which causes a misalignment that contributes to failed grasps at high speeds.

\section{Discussion}
\label{sec:discussion}

We advance the state of the art towards deployable, aggressive, and versatile aerial manipulation.
We accomplish this by combining the desirable properties of a soft gripper with an advanced quadrotor platform and a state-of-the-art perception system.
The gripper's reliable performance ---even with imperfect grasp placement--- and its ability to decouple the quadrotor dynamics from contact constraints enable high speed grasps. 
The system's ability to operate outside of a controlled motion-capture room is vital in nearly every practical application, enabling aerial vehicles to perform manipulation tasks such as emergency supply distribution, package pickup and delivery, and warehouse automation.

The compliant fingers of our soft gripper are key to enabling the full system to grasp at high speeds.
As the drone flies faster, its position estimate suffers and tracking errors increase.
Furthermore, the detections of the target's pose have several centimeters of error and VIO drift adds additional error.
In spite of these error sources, the system successfully picks up objects at over 2~m/s, with full onboard vision. 

The gripper mechanism is mechanically simple, which we have found to be a great advantage in manufacturability and repairability.
Each finger assembly requires only a handful of 3D-printed parts, fishing line, off-the-shelf foam, and a small motor.
The fingers themselves have been surprisingly resilient when attached to the drone.
Their most common failure modes are tears that develop near the base, which we have been able to easily repair with common super glue.
They also provide cushion to the drone during hard landings;
during our development we have seen the drone fall over six feet onto solid surfaces and not require any repair.
Even in these cases, the fingers did not break and were used successfully for subsequent grasps.

Controlling soft fingers can be challenging, as they are difficult to model.
In comparison to rigid manipulators, which have a well-defined finite dimensional state space, soft fingers are a continuum and have an infinite-dimensional state space.
This makes it difficult to relate control input to finger state, which is necessary for our gripper to open its fingers for object grasping while staying out of the cameras' field of view when necessary.
We have demonstrated that FEM-based modeling and control enable us to accurately approximate the  continuum mechanics of our soft fingers, enabling reliable grasp performance. Our approach is readily generalizable to different, task-specific gripper configurations, which could be designed to exploit the morphology of a specific target, following the geometric analysis in Supplementary Text S1.

We see several potential avenues for future work. One limitation of our pose estimation pipeline is its assumption of prior knowledge of the target object's geometry and visual features.
Leveraging recent advances in category-level keypoint detection, e.g.,~\cite{manuelli2019kpam}, would enable the system's perception pipeline to generalize to other instances of the same semantic category (e.g., multiple types of packages or soda bottles).
Our drone is also currently limited to picking up light objects.
We expect that scaling up the drone platform relative to the gripper would lead to a wider range of graspable objects without having to change the rest of the system, as a larger drone would have more control authority and more reliable trajectory tracking both before and after grasp.
Finally, full vision-based grasps of moving targets is an immediate next step for our system.
Grasping a moving target using only vision-based inputs with our current system requires predicting the object's motion when it goes out of frame before the grasp point, inevitably adding error to the object pose estimate.
Timing offsets between the drone's pose estimate and the camera image frame  also manifest as error in the object pose estimate.
These two additional sources of error currently prevent our drone from performing vision-based grasps of moving targets.
More sophisticated trajectory prediction techniques or constraints on the target's possible motion might be needed to enable these moving grasps.

\section{Materials and Methods}
\label{sec:materialsmethods}

\subsection{Hardware Design}
\subsubsection{Drone Design}
Our drone platform (\cref{fig:overview}C) is a quadrotor inspired by the \emph{Intel Aero Ready to Fly} (RTF) drone. It has a custom waterjet carbon fiber frame, but uses the same Yuneec brushless DC motors and propellers as the RTF drone. A Pixhawk 4 Mini flight controller running custom PX4 firmware handles the low-level adaptive controller. The drone is powered by a four-cell Lithium Polymer battery, which provides about three minutes of flight time (enough for three grasp trajectories).

The drone has a front-facing Realsense D455 RGB-D camera pitched down at a 35 degree angle for target pose estimation, and a rear-facing Realsense T265 for VIO estimation.
We observed the T265's odometry estimation failing under high-vibration conditions, so we added vibration damping silicone grommets between the camera and drone. 
The camera required physical masking on the bottom of the camera lens to match the field of view constraint used in the gripper's tendon length optimization, as the camera does not support software-defined region masking.
Ensuring that the fingers were not in the camera frame was vital, as the visual odometry estimate when the fingers entered the frame proved to be quite poor.

The visual target estimation pipeline and trajectory planner are run on a Jetson Xavier NX. Both Realsense cameras are connected to the Xavier via USB 3.0 and a powered USB hub. The Xavier communicates with the Pixhawk over UART. When using motion-capture pose estimates instead of visual navigation, the Xavier receives pose updates over Wifi from a base station connected to the motion capture system. See Table S4 for detailed component specifications.

\subsubsection{Gripper Design}
The drone interfaces with the gripper through standoffs connected to a 3D printed base plate, which houses the gripper components (\cref{fig:overview}D). A custom-made, lightweight PCB embeds an ATMega32U4 microcontroller to handle logic control, 2 DRV8434 motor controllers which actuate four 31:1 gear ratio 12V DC motors, and a voltage regulator and connector to convert the Lipo battery's voltage into a stable 12V. Jumper cables are wired from GPIO pins on the Xavier to the PCB to connect the gripper state machine with the low level gripper controller.

The gripper features four passively closing, cable actuated fingers (\cref{fig:fem}B). Each finger is molded using Smooth-On \textit{FlexFoam-iT! X}~\cite{flexitfoam} and a custom 3D-printed mold, which follows a circular arc (Fig.~S4). The initial state of the finger is then its closed position, which is maintained by the elasticity of the mesh. A 3D-printed finger-base connector is super glued to the top of the finger and can be slotted into the base plate for quick replacement. The tendon guides are positioned between evenly placed foam extrusions and adhered with super glue. Braided fishing line is fixed to the last tendon guide and the motorized winch, passing through the interior tendon guides. As the winch tightens, the tendon contracts, compressing the outer edge of the finger and forcing it open. As the winch loosens, the elastic force of the mesh returns the finger to its default closed position.

\subsection{Target Object Perception and Pose Estimation}

We present a robust pose estimation system that combines a keypoint detector with a state-of-the-art algorithm 
 for kepoint-based 3D pose estimation, namely Teaser++~\cite{Yang20tro-teaser}. %
Estimated poses are further refined through a fixed-lag smoother (\cref{fig:flow_chart}). We also developed a semi-automated data annotation tool to rapidly train the keypoint detector.

\subsubsection{Semi-Automated Keypoint Annotation}
Keypoint annotation can be a costly process, requiring multiple points to be labeled in each image of a large image dataset (\eg with tens of thousands of images).
We draw inspiration from prior methods of speeding up keypoint labeling by using known camera poses at training time~\cite{blomqvist2022semiautomatic,liu2020keypose}, enabling a small number of manually labeled keypoints to be projected into many images. 
Fig.~\ref{fig:pose_plots}A presents an overview of the annotation pipeline.
The target object and calibration board are positioned such that both are visible in the recorded camera frames, and they are fixed for the entirety of the recording process.
We move the D455 camera at varying distances and angles from the object and record all RGB and depth images.
The pose of the camera with respect to the calibration board for each image {can be readily determined by computing the camera extrinsics using the calibration board}.
Known camera poses enable reasoning about the 3D position of each keypoint across the whole trajectory sequence.
First, pixel locations for the keypoints in user-selected images are manually annotated using the Matlab tool GUI (Fig.~S5).
The 3D position for each keypoint relative to the camera is determined by back-projecting the pixel coordinates using the depth of each keypoint, given by the D455.
The keypoint positions can be further refined by minimizing the reprojection error in a small number of manually labeled training images, and then reprojected into all other images based on the camera model.

In our tests, we only annotate keypoints on the top face of the object and assume that no keypoints are occluded in our training set; this is a valid assumption considering the viewing angle of our drone.
Proper annotation of keypoint visibility is important to prevent the neural net from associating features blocking the view of the keypoint with the true keypoint features.
The labeling method does support keypoints placed on all sides of the object, but self-occlusions require visibility detection. This can be done by sampling the depth of each keypoint using the depth image and comparing that against the expected depth obtained by transforming the keypoint relative to the camera. 
If there is a large discrepancy, the keypoint is assumed to be occluded.
This method requires a highly accurate depth sensor and the D455 proved to be inadequate.

We add variety in the training set to improve generalization of the keypoint detector by collecting data in several environments: a motion capture room, a cluttered workshop, a lounge space, a foyer, and an outdoor field (Fig.~S6).
For each location, the data collection process was repeated multiple times with different poses of the target object.
A randomly selected subset of the runs were added to the validation set.
Across the three targets, 39548 training images and 15172 validation images were collected and annotated, within only a few days of manual labor.

\subsubsection{Keypoint Detection}

To minimize training time and increase generalizability, we apply transfer learning on a state-of-the-art keypoint detection architecture, \emph{trt-pose}~\cite{trtpose}.
The network is comprised of a ResNet-18 backbone, followed by a keypoint prediction head.
The training data is augmented using standard techniques (perturbations of translation, rotation, scale, and contrast) to further improve diversity of the training set. We present the full set of parameters we used for trt-pose's training pipeline in Table S5.
Training takes around seven hours on an Nvidia GeForce GTX 1080 Ti GPU, with the loss converging after 75 epochs.

The original RGB images have size ${1280\times720}$ and are cropped by removing the top half and resized into $912 \times 256$ images to improve inference speed. 
Due to the geometry of the camera mount, cropping the top half of the images does not reduce performance as objects in that region are too far away to detect reliably anyway.
The trained network is compiled to leverage the Jetson Xavier's hardware using TensorRT, and runs at approximately 14 Hz on the Xavier.

Fig.~\ref{fig:pose_plots}D shows the results of the keypoint detector on the \medkit test set, along with visual representations in Fig.~\ref{fig:pose_plots}C.
Keypoint error is defined as the pixel distance between the estimated and ground-truth keypoints, averaged across all keypoints, and only for keypoints that were valid detections (an invalid detection occurs when the keypoint isn't detected and the network predicts $(0,0)$).
The error remains roughly constant with distance, but we note that pixel error is influenced by the scale of the target.
At far distances, the target's perceived area is smaller and a unit of pixel error translates to higher error in world coordinates than a unit of pixel error at close distances.

\subsubsection{Robust Point Cloud Registration}
From the depth image provided by the D455 camera and the detected keypoint pixel coordinates, we create an estimated point cloud of keypoints, denoted as $\mathcal{A} = \{\bm{a}_i\}_{i=1}^N$.
Furthermore, the corresponding keypoints are also annotated on the known object CAD model through the data annotation tool, leading to a second point cloud, namely $\mathcal{B} = \{\bm{b}_i\}_{i=1}^N$.
Solving for the transformation between these two point clouds allows us to estimate the pose of the target relative to the camera.
We seek to find the rotation $\bm{R} \in SO(3)$ and translation $\bm{t} \in \mathbb{R}^3$ which minimize the following \emph{truncated nonlinear least squares} problem~\cite{Yang20tro-teaser}:
\begin{equation}
    \min_{\bm{R}\in SO(3), \bm{t} \in \mathbb{R}^3} \sum_{i=1}^{N} \min \left(|| \bm{b}_i - \bm{R} \bm{a}_i - \bm{t}||^2 \;,\; \bar{c}^2 \right),
    \label{eq:TLS}
\end{equation}
where $\bar{c}$ represents the upper bound on the residual error for a measurement to be considered an inlier, and is set to be 10~mm.
In the truncated least squares problem in Eq.~\ref{eq:TLS}, 
measurements with residual error smaller than $\bar{c}$ are used to compute a least-squares estimate, while residuals greater than $\bar{c}$ have no effect on the estimate and are discarded as outliers. In our application, this is useful for filtering out outliers caused by spurious keypoint or depth detections.
In practice, we utilize Teaser++~\cite{Yang20tro-teaser} to solve this problem, which takes on average 2 ms to solve a single registration problem on the Xavier NX.
Fig.~\ref{fig:pose_plots}E-F demonstrate the results of our approach for the \medkit object.
Translation error is defined as the Euclidean distance between the estimated and ground-truth translations, while rotation error is computed as the geodesic distance between the estimated and ground-truth rotations. Additionally, we only report data for detections that had at least one inlier point.
We nominally plan trajectories at a distance of 1.25~m from the target, which equates to roughly 5~cm of translation error and 15~degrees of rotation error.
Our experiments have validated that this error is well tolerated when combined with the fixed-lag smoother and the robustness of the soft gripper.

\subsubsection{Fixed-Lag Smoother}

Estimates of the target's relative pose from individual images are fused together in a fixed-lag smoother to reduce noise in the estimate and provide estimates for linear and angular velocity.
Knowledge of the target's velocity is necessary for grasping moving targets at the desired relative speed. 
We rely on the estimation process described below to infer velocity.

We model the target's motion as a nearly constant velocity model~\cite{BarShalom01}, with the target's pose $\bm{T}_k \in SE(3)$ and linear and angular velocity $\bm{\xi}_k \in \mathbb{R}^6$ at time $k$: %
\begin{equation}
\begin{split}
&\bm{T}_{k+1} = \bm{T}_{k} \boxplus \left(\bm{\xi}_{k} dt\right)\\ %
&\bm{\xi}_{k+1} = \bm{\xi}_{k} + \bm{\nu}_{k}dt,
\end{split}
\end{equation}
where $\bm{\nu}_k \in \mathbb{R}^6 \sim \mathcal{N}(\bm{0}, \Sigma)$ represents white-noise angular and linear acceleration, 
and $\boxplus$ maps a vector in $\mathbb{R}^6$ (on the right-hand-side of the operator) to $SE(3)$ (e.g., via an exponential map or, more generally, a \emph{retraction}~\cite{Forster17tro}) and then composes it with the pose on the left-hand-side of the operator.
Given observations of the target's pose over the last $\tau$ timesteps, namely $\hat{\bm{T}}_{T-\tau:T}$, we estimate $\bm{T}_{T-\tau:T}$ and $\bm{\xi}_{T-\tau:T}$ as
\begin{equation}\label{eq:fixed_lag_smoother}
\begin{split}
\min_{\bm{T}_{T-\tau:t_T}, \bm{\xi}_{T-\tau:T}} &\sum_{k=T-\tau}^{T} ||\bm{T}_k~\boxminus~\hat{\bm{T}}_k||_{\Omega_1}^2\\
&+ || (\bm{T}_{k}~\boxplus~(\bm{\xi}_{k}dt))~\boxminus~\bm{T}_{k+1} ||_{\Omega_2}^2\\
&+ || \bm{\xi}_k - \bm{\xi}_{k+1} ||_{\Omega_3}^2 + ||\bm{\xi}_k||_{\Omega_4}^2,
\end{split}
\end{equation}
where %
$\boxminus$ denotes a tangent space representation of the relative pose between two poses. 
The first term encourages the estimate to match the target pose observations, the second term constrains the estimated velocity to be consistent with the poses, and the third term forces changes in velocity between timesteps to be small. The final term provides regularization on the velocity estimate. We found a small amount of velocity regularization to be necessary for consistent initialization of the filter. $\Omega_1,~\Omega_2,~\Omega_3,~\Omega_4$ are tunable weight matrices to trade off among the four terms (see Table S6). The optimization is implemented in C++ using the GTSAM library~\cite{gtsam_github}. The incremental iSAM2 algorithm~\cite{Kaess12ijrr} is used to optimize \eqref{eq:fixed_lag_smoother}, as it enables reuse of computation between timesteps.

\subsection{Trajectory Optimization and Control}

To enable our grasps,
we compute a smooth polynomial trajectory for the drone, passing at a set distance over the target (the \emph{grasp point}).
This nominal trajectory is tracked with an adaptive controller which compensates for the added mass after grasping and for other external disturbances.
Throughout the grasp procedure, the drone's current position is fed into a gripper state machine, which sets the finger winch angle based on the results of an offline FEM-based optimization (\cref{fig:flow_chart}).
The trajectory optimization and control are based on prior implementations for motion-capture based grasping~\cite{Fishman21iros-softDrone2}, with added functionality enabling trajectory planning during flight and updating of the reference trajectory to account for moving targets.

\subsubsection{Minimum-Snap Polynomial Trajectory Optimization}
We compute a fixed-time polynomial aligned with the target's grasp axis and constrained to start at the drone's initial position, pass through the grasp point, and end at an arbitrary final position.
The grasp point is defined as the target's position, plus a tunable offset to account for the length of the fingers and any misalignment between the target frame center and the geometric center.
We seek to find a polynomial that minimizes the 4th derivative of position, or snap, of the drone.
As shown in~\cite{Mellinger11icra}, polynomials of this form enable smooth, continuous motion which allows for stable, consistent grasps.
The key idea which enables grasping moving targets is that the polynomial is planned with respect to the \textit{target's} frame, rather than the inertial frame.
As the target moves the polynomial moves with it, inducing a disturbance which is compensated for by our low-level tracking controller.
The polynomial provides a series of position, velocity, and acceleration setpoints that are tracked by the drone's flight controller.
The drone's yaw setpoint is set to always point toward the estimated object position.
Supplementary Text S2 provides further exposition on our specific trajectory optimization implementation.

\subsubsection{Adaptive Quadrotor Control}
The drone is subject to various forms of disturbances that could impede both grasp success and post-grasp flight performance.
After the target is grasped, the effective mass of the quadrotor increases and the thrust efficiency decreases, as the grasped object partially obstructs the propellers.
Estimates of system model parameters are also imperfect, and benefit from online adaptation.
Additionally, we observe that our system is also sensitive to variable disturbances such as wind in outdoor environments or the ground effect.
To mitigate these disturbances, we adopt the geometric adaptive control law presented in~\cite{Goodarzi15adaptive}, that we briefly review below.

 The external disturbances in our system primarily affect the translational dynamics of the drone, with negligible impact on the rotational dynamics. Hence, we model the quadrotor dynamics as: 
 \begin{equation}
 \begin{array}{ll}
     m \Ddot{\quadpos} = m \vg + f \rotcolz + \disturbance_f \\
     \dot{\quadrot} = \quadrot \hat{\quadrvel} \\
     \MJ \dot{\quadrvel} = - \quadrvel \times \MJ \quadrvel + \vtau \\
 \end{array}
 \label{eq:quadrotor_dynamics}
 \end{equation}
where $m$ is the drone's mass, $\MJ$ is the moment of inertia, $\vg$ is the gravity vector, $\quadpos \in \Real{3}$, $\dot{\quadpos} \in \Real{3}$, $\quadrot \in \SOthree$, $\quadrvel \in \Real{3}$ are the position, velocity, rotation, and angular velocity of the drone, respectively, $f$ is the scalar thrust force applied along the local vertical direction $\rotcolz$, $\vtau$ is the torque applied by the quadrotor, and $\disturbance_f$ is the unknown translational disturbance. The 
\emph{hat} operator $\hat{\cdot}$  maps a vector in $\mathbb{R}^3$ to an element of the Lie algebra $\mathfrak{so}(3)$, i.e., a $3 \times 3$ skew symmetric matrix; see~\cite{Goodarzi15adaptive}.

 We then compute the quadrotor's thrust and moment as:
 \bea
 \label{eq:adaptiveGeoController}
  f &=& -\rotcolz\tran ( k_p \ve_p + k_v \ve_v+ 
                         m \vg - m \Ddot{\quadpos}_d + 
                         \estDisturbance_f)  \\
        \vtau &=& -k_r \ve_r - k_{\Omega} \ve_{\Omega} + \MJ \quadrot \tran \quadrot_d \dot{\quadrvel}_d \nonumber \\
        & & + (\quadrot\tran \quadrot_d \quadrvel_d)^{\wedge{}} \MJ \quadrot\tran \quadrot_d \quadrvel_d 
 \eea
 where $\quadpos_d$, $\quadrot_d$, $\quadrvel_d$ are the desired position, attitude, and angular velocity, respectively,
 $\ve_p$, $\ve_v$, $\ve_r$, $\ve_{\Omega}$ are the position, velocity, rotation, and angular velocity errors, 
 $\estDisturbance_f$ is the estimated disturbance on the translation dynamics,
 and $k_p,k_v,k_r,k_{\Omega}$ are user-specified control gains. 
 We refer the reader to~\cite{Goodarzi15adaptive} for the definition of the errors and a more comprehensive discussion.

 The estimated disturbance is adjusted online using the following law:
 \bea
 \label{eq:adaptiveLaw}
 \frac{d\estDisturbance_f}{dt}\!=\! \Pi\left( \gamma_f (\ve_v \!+\! k_{af} \ve_p) \right)
 \eea
 where $\gamma_f$, $k_{af}$ are user-specified gains
 and $\Pi$ is a suitable projection function.

This formulation ensures that the tracking errors asymptotically go to zero even in the presence of unknown disturbances~\cite{Goodarzi15adaptive}.

\subsubsection{FEM-Based Gripper Control}
To optimally actuate the gripper, the controller must deform its fingers such that they effectively enclose the target object, while simultaneously ensuring that they do not obstruct the view of the onboard cameras.
To balance these two goals, we employ an optimization-based control method. %
At the core of our control method is a finite element-based (FEM) model.

The gripper's volume is discretized into a finite element mesh with nodal positions $\bm{x}$, and simulated cables are routed through this mesh.
Given cable contractions $\bm{u}$, we can solve for a corresponding statically stable pose
\begin{equation}
    \label{eq:fem-physics}
    \bm{x}(\bm{u}) = \arg\min_{\bm{x}} E(\bm{u}, \bm{x})
\end{equation}
by minimizing the total potential energy of the system $E$ using Newton's method.
This general approach can be elegantly extended to account for dynamics, and is done in~\cite{Bern19rss,li2020incremental}.

Our control optimization is a \emph{nested optimization}.
Its objective $\mathcal{O}(\bm{u}, \bm{x}(\bm{u}))$ is written in terms of the physically valid pose $\bm{x}(\bm{u})$, found by solving Eq.~\ref{eq:fem-physics}.
We minimize this objective using gradient descent to find optimal control inputs for a desired configuration
\begin{equation}
\bm{u}^* = \arg\min_{\bm{u}}\mathcal{O}(\bm{u}, \bm{x}(\bm{u})).
\end{equation}
Computing the gradient of our control objective requires the Jacobian $\frac{d\bm{x}}{d\bm{u}}$ relating changes in control inputs $\bm{u}$ to changes in the corresponding physically-valid mesh shape $\bm{x}(\bm{u})$, which we solve for using direct sensitivity analysis, as described in \cite{bern2021soft}.

Our specific control objective $\mathcal{O}$
is the weighted sum of two sub-objectives, which are illustrated in Fig.~\ref{fig:fem}D-E:
\begin{equation}
    \mathcal{O}= \mathcal{O}_{\text{grasp}} + \mathcal{O}_{\text{FOV}}
\end{equation}

The first sub-objective, $\mathcal{O}_{\text{grasp}}$, drives the gripper's fingers towards a configuration where they can effectively grasp an object.
The sub-objective
\begin{equation}
    \mathcal{O}_{\text{grasp}} = \left|\left|(\bm{s} - \bm{y}_1(\bm{u})) \times (\bm{s} - \bm{y}_2(\bm{u}))\right|\right|^2
\end{equation}
works by maximizing the area enclosed by the gripper's fingertips and a \emph{target point} $\bm{s}$ (\cref{fig:fem}E, \cite{Fishman21iros-softDrone2}). Here, $\bm{s}$ represents the centroid of the target shifted from the drone by a nominal offset which approximates the relative position between the drone and target at the moment the gripper begins to close.
The fingertip positions $\bm{y}_1(\bm{u})$ and $\bm{y}_2(\bm{u})$ are defined in terms of the physically-valid mesh position $\bm{x}(\bm{u})$ using barycentric coordinates.

The second sub-objective, $\mathcal{O}_{\text{FOV}}$, ensures the the gripper's fingers do not obstruct the camera's field of view.
This is accomplished by defining a plane with point $\bm{p}$ and normal $\bm{n}$, {analogous to the bottom face of a camera's view frustum} (\cref{fig:fem}D).
To prevent the finger from entering into the camera's field of view, the sub-objective
\begin{equation}
    \mathcal{O}_{\text{FOV}} = \sum_{i}f\left((\bm{p} - \bm{x}_i(\bm{u})) \cdot \bm{n}\right)
\end{equation}
sums over the squared penetration depths of all nodes in the mesh.
Here, the function
\begin{equation}
    f(z) =
    \begin{cases}
    z^2 - \varepsilon z + \frac{\varepsilon^2}{3}, & \text{if $\varepsilon < z$,} \\
    \frac{z^3}{3\varepsilon}, & \text{if $0 < z \leq \varepsilon$,} \\
    0, & \text{otherwise}
    \end{cases}
\end{equation}
is a smooth one-sided quadratic, driving the mesh outside the camera's field of view while ensuring the optimizer is well-behaved. Depending on the grasp phase, $\mathcal{O}_{\text{FOV}}$ is added for both the front and the rear camera (to obtain the \emph{target observation state}) or only for the rear camera (to obtain the \textit{pre-grasp state}).

\subsection{Measuring Sources of Error}
\label{sec:error_sources}
The success of our vision system depends on three sources of error: target estimation error, VIO drift, and trajectory tracking error.
We collect motion capture data alongside the vision flights to benchmark these errors, although the quadrotor only uses vision-based data for its own control.

To quantify the amount of target estimation error (\cref{fig:vision_static}B), we placed motion capture markers near the target, but far from the key visual features.
The target's motion capture frame was manually adjusted until it visually appeared aligned with the expected target's vision frame.
Estimation error is then reported as the difference between our vision based estimate and the motion capture ground truth.
VIO drift (\cref{fig:vision_static}C) was measured by aligning the grasp trajectory as reported by the vision system with the recorded motion capture data.
This alignment was done using \emph{evo}~\cite{Grupp17evo}, which performs a least squares optimization to find the optimal rigid transformation between trajectories.
Here, we are only concerned with aligning part of the trajectory up until the grasp point, as post-grasp drift generally has little effect on grasp success.
Therefore, we align trajectories starting from 9 seconds before the grasp time up until the grasp time.
We found this range to be suitable to be sure that the resulting alignment errors were due to trajectory drift and not because of a lack of data points.
Trajectory tracking errors (\cref{fig:vision_static}D-G) were analyzed by extracting the setpoint and pose estimate from the flight controller.

\section*{Acknowledgments}
We would like to thank M. Mohamoud for initial quadrotor design, B. Evans for work assembling and manufacturing the gripper electrical and mechanical components, W. Menken and N. Mondal for electrical design of the gripper circuit board, and J. Fishman for useful discussion on soft aerial manipulation. We would also like to thank V. Murali, P. Lusk, and N. Hughes for feedback on initial drone design and system troubleshooting.

\textbf{Funding:} This work has been partially sponsored by an MIT Research Support Committee (RSC) award, Carlone's Amazon Research Award, and by MathWorks. In addition, A.R. was supported by a National Defense Science and Engineering Graduate Fellowship. 

\bibliography{../../references/refs,../../references/myRefs.bib}
\bibliographystyle{IEEEtran}

\clearpage
\section*{Supplementary Materials}

\captionsetup[table]{labelformat=empty}
\captionsetup[figure]{labelformat=empty}

 \begin{figure*}[h]
    \centering
    \includegraphics[width=\textwidth]{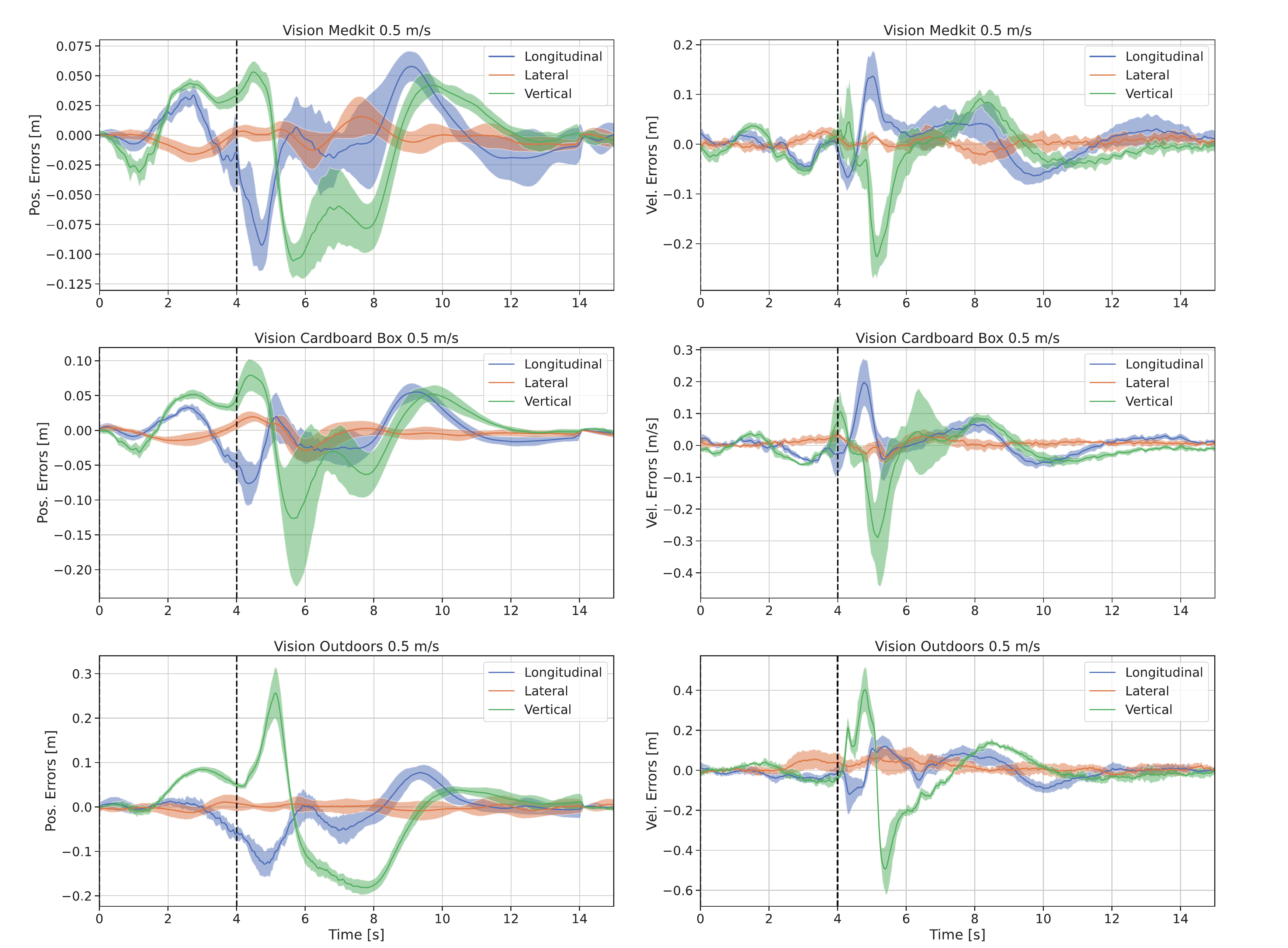}
     \caption{\textbf{Figure S1. Tracking error breakdown for experiments with static target  (continued below).} Longitudinal, lateral, and vertical position and velocity tracking errors for the vision-based experiments with static target.  Before grasp, vertical errors become more positive due to ground effect. Post-grasp, the mass of the object results in more vertical errors, that are gradually compensated for.}
\end{figure*}

 \begin{figure*}[h]
    \centering
    \includegraphics[width=\textwidth]{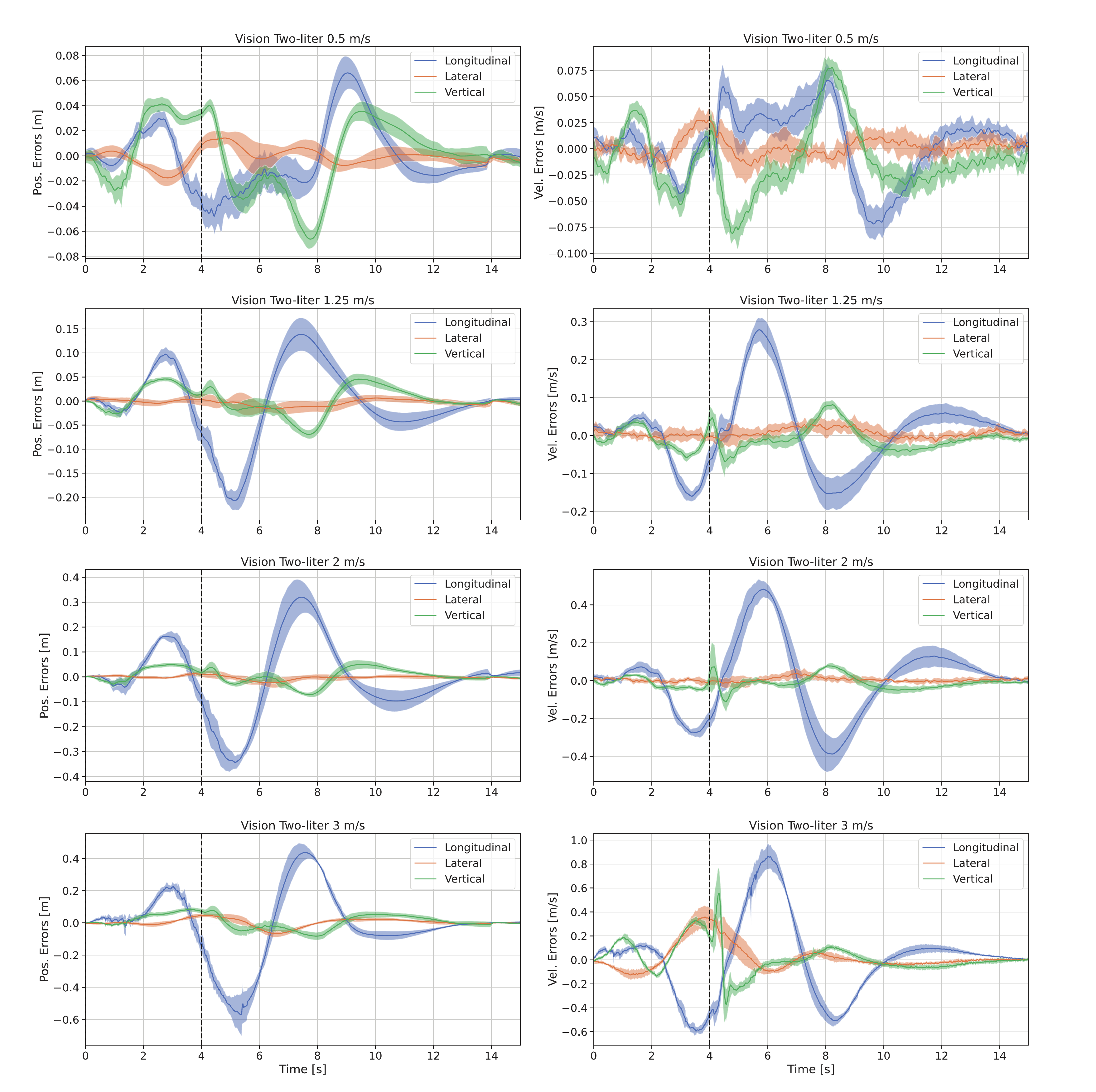}
     \caption{\textbf{Figure S1. Tracking error breakdown for experiments with static target (continued).} 
 As the velocity increases, we see that longitudinal errors dominate while lateral and vertical errors increase slightly. Large longitudinal tracking errors have little effect on grasp success, as the timing of the gripper closure is based on the drone's actual position with respect to the target. It is important, however, that the lateral and vertical errors are minimal at the time of grasp closure. \label{fig:appdx_static}
     }
\end{figure*}

 \begin{figure*}[h]
    \centering
    \includegraphics[width=0.93\textwidth]{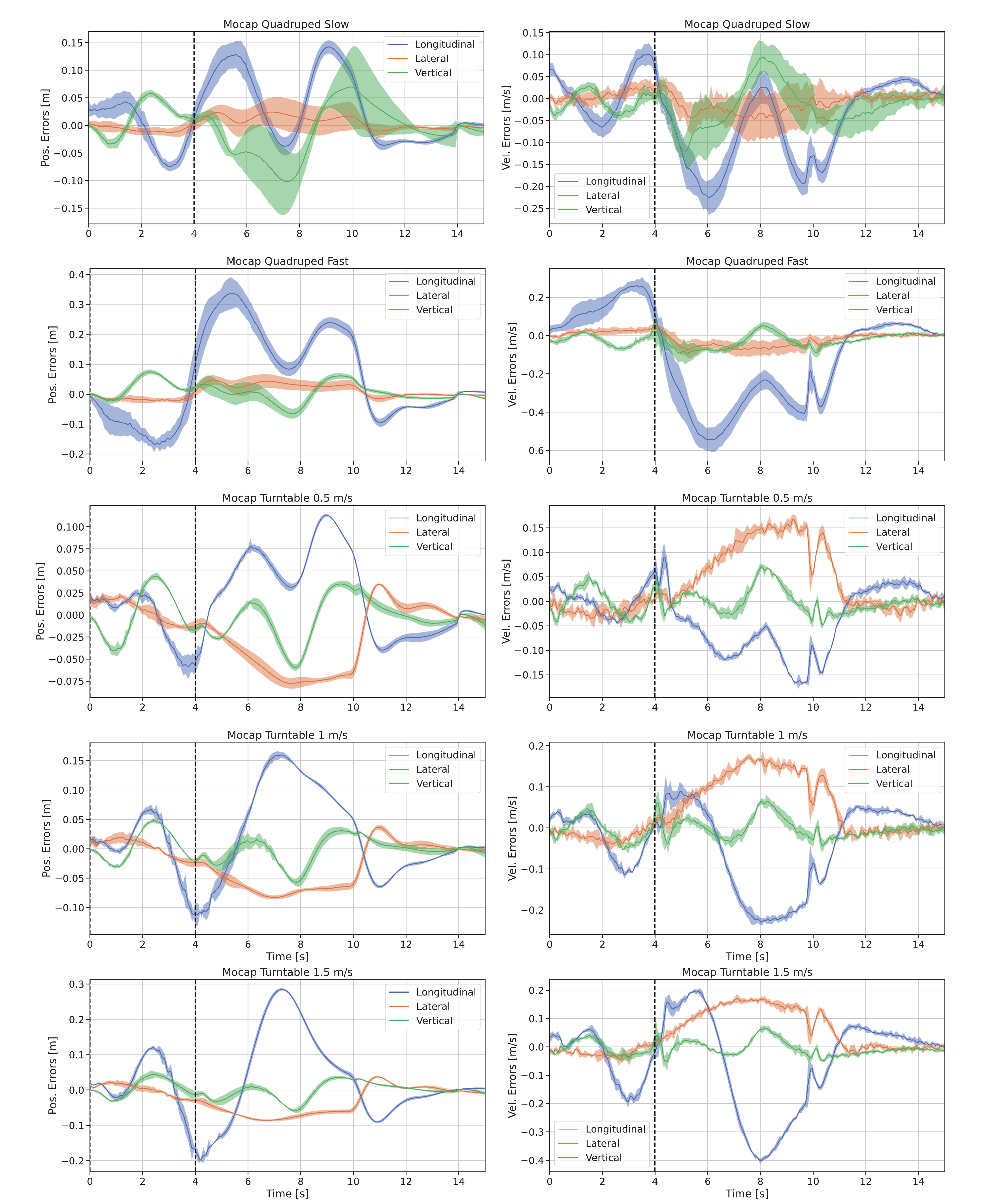}
     \caption{\textbf{Figure S2. Tracking error breakdown for experiments with moving target.} Longitudinal, lateral, and vertical position and velocity tracking errors for the motion-capture-based experiments with moving targets. For the quadruped experiments, longitudinal errors increase greatly with increased linear speed of the target. Lateral and vertical errors also see slight increases with speed, due to either the higher variation of the target's motion or worsening of the drone's control authority at this speed. We see that there is increased lateral errors when grasping a rotating target, due to the desired grasp direction also rotating. These errors increase with increased relative grasp velocity. }\label{fig:appdx_moving}
\end{figure*}

 \begin{figure*}[h]
    \centering
    \includegraphics[width=\textwidth]{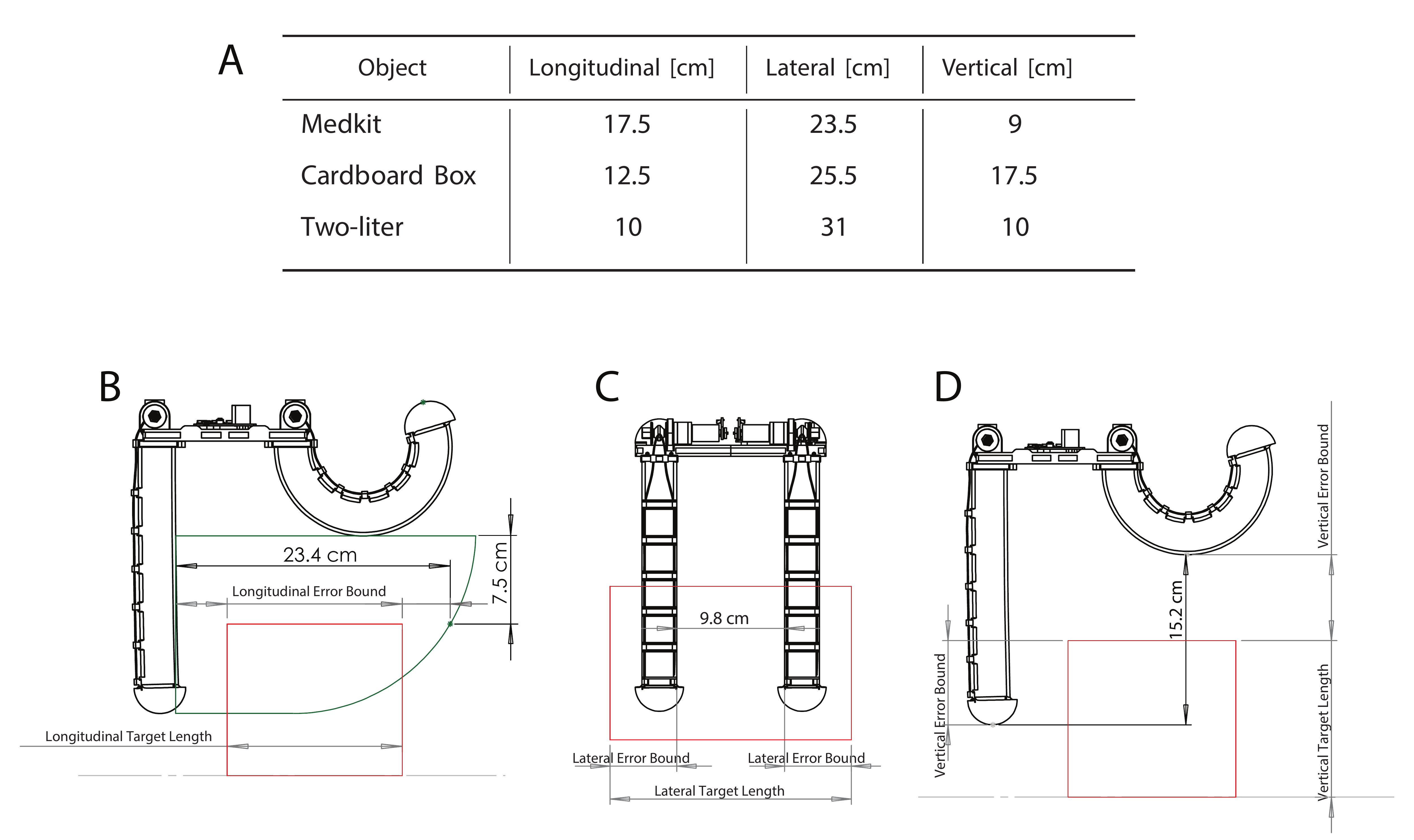}
     \caption{\textbf{Figure S3. Gripper dimensions and error bounds.} \textbf{(A)} The table shows the dimensions of the three objects. \textbf{(B)} The longitudinal error bound. Because the fingers close in roughly a circular arc, the longitudinal gripper length decreases as it closes. Assuming the lowest point of the front finger is 7.5 cm above the top face of the target, our effective longitudinal gripper length can be computed when the finger reaches the same height as the target's top face. This effective gripper length minus the target's longitudinal length, halved, gives us our error bound.   \textbf{(C)} The lateral error bound. If the drone has greater lateral error than this bound, then one half of the gripper will not make contact with the target. \textbf{(D)} The vertical error bound. If there exists high error pointing downward, the front fingers will collide with target. Conversely, if there is high error pointing upward, the rear fingers will not make contact.} \label{fig:appdx_moving}
\end{figure*}

 \begin{figure*}[h]
    \centering
    \includegraphics[width=\textwidth]{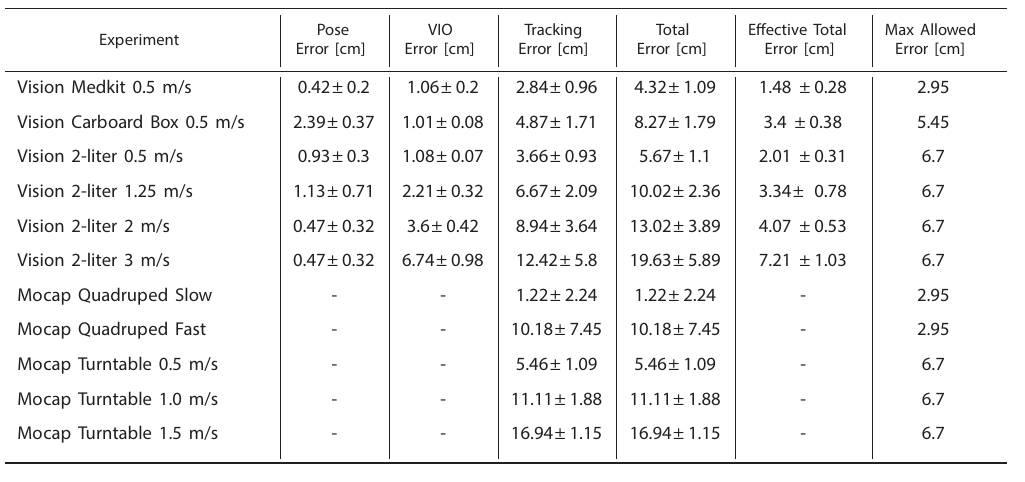}
     \caption{\textbf{Table S1. Longitudinal errors at the time of grasp.} Longitudinal pose estimate, VIO, tracking errors, total error, and maximum allowed error for each experiment. As the majority of the grasp's motion happens in this axis, we see much higher tracking errors compared to the lateral and vertical axes. These errors increase greatly with increased velocity, due to limited control authority. However, these tracking errors have little effect on the success of the grasp. If the drone is aligned vertically and laterally, the grasp will succeed if it is triggered at the proper time, which is affected by VIO drift and our initial pose estimate. We can then compute our effective total error, which is the sum of only these two sources. We see that in the majority of our experiments, the effective total error remains within the maximum allowed longitudinal error bound derived in Supplementary Text S1.}\label{fig:longitudinal}
\end{figure*}

 \begin{figure*}[h]
    \centering
    \includegraphics[width=\textwidth]{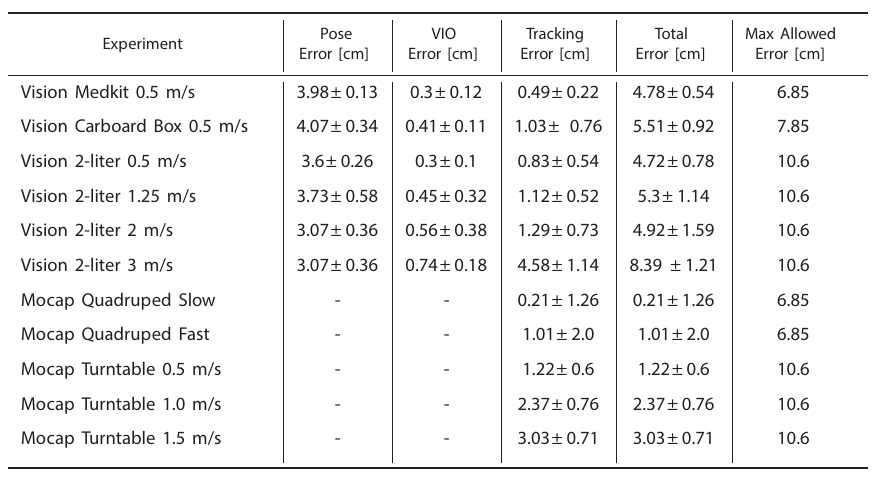}
     \caption{\textbf{Table S2. Lateral errors at the time of grasp.} Lateral pose estimate, VIO, tracking errors, total error, and maximum allowed error for each experiment. We see high lateral error in our pose estimate, which we note could be accounted for by alignment errors in the ground truth pose. VIO is minimal for all speeds, while tracking errors increase with speed and are effected by the rotational motion of the turntable experiments. For all experiments, 
   the total lateral error remains within the maximum allowed lateral error bound derived in Supplementary Text S1. Despite this, we observe complex grasp interactions such as buckling or twisting of the fingers which result in fingers failing to make secure contact in the lateral direction, and, consequently, failed grasps.}\label{fig:lateral}
\end{figure*}

 \begin{figure*}[h]
    \centering
    \includegraphics[width=\textwidth]{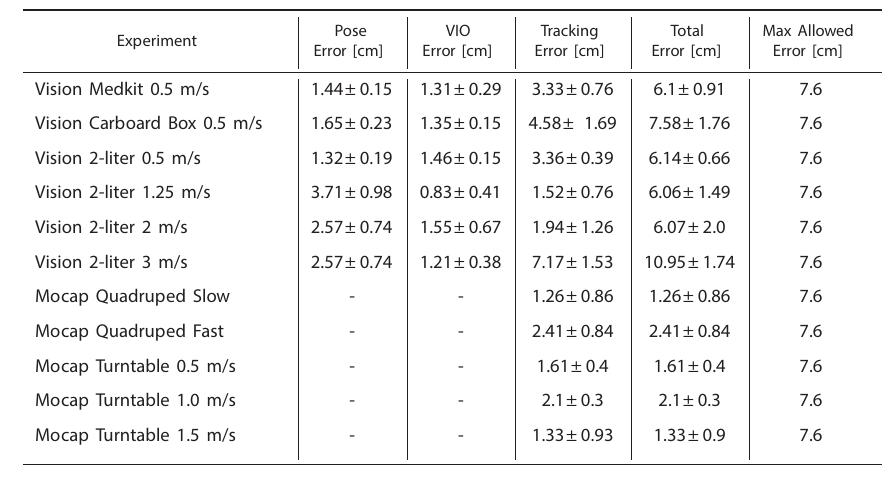}
     \caption{\textbf{Table S3. Vertical errors at the time of grasp.} Vertical pose estimate, VIO, tracking errors, total error, and maximum allowed error for each experiment. We see that pose and VIO errors are minimal for all experiments. Differences in the pose estimate for the two-liter experiments are explained by slight alignment differences when defining the ground truth, motion capture pose for the bottle. Tracking errors at the point of grasp are explained by ground effect, unintended pre-grasp interaction with the target, and general control authority limitations. In particular, we see high vertical tracking error at the fastest speed. This could be due to the coupling of forward velocity and altitude control for quadrotors, where our altitude authority worsens at higher speeds. The remaining experiments remain within our vertical error bounds (derived in Supplementary Text S1), albeit on the boundary in some cases. Indeed, vertical misalignment is a common failure mode. }\label{fig:vertical}
\end{figure*}

\begin{figure*}[h]
    \centering
    \includegraphics[width=\textwidth]{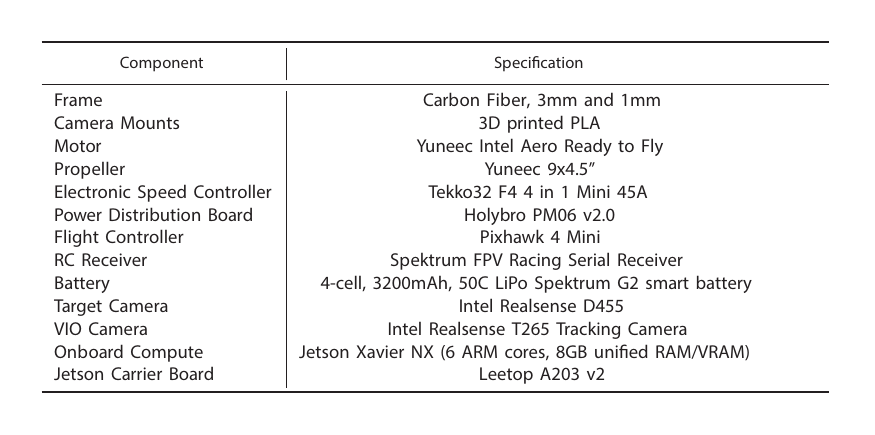}
     \caption{\textbf{Table S4. Drone specifications.} Component list for the drone platform and the perception system.}\label{drone_details}
\end{figure*}

 \begin{figure*}[h]
    \centering
    \includegraphics[width=\textwidth]{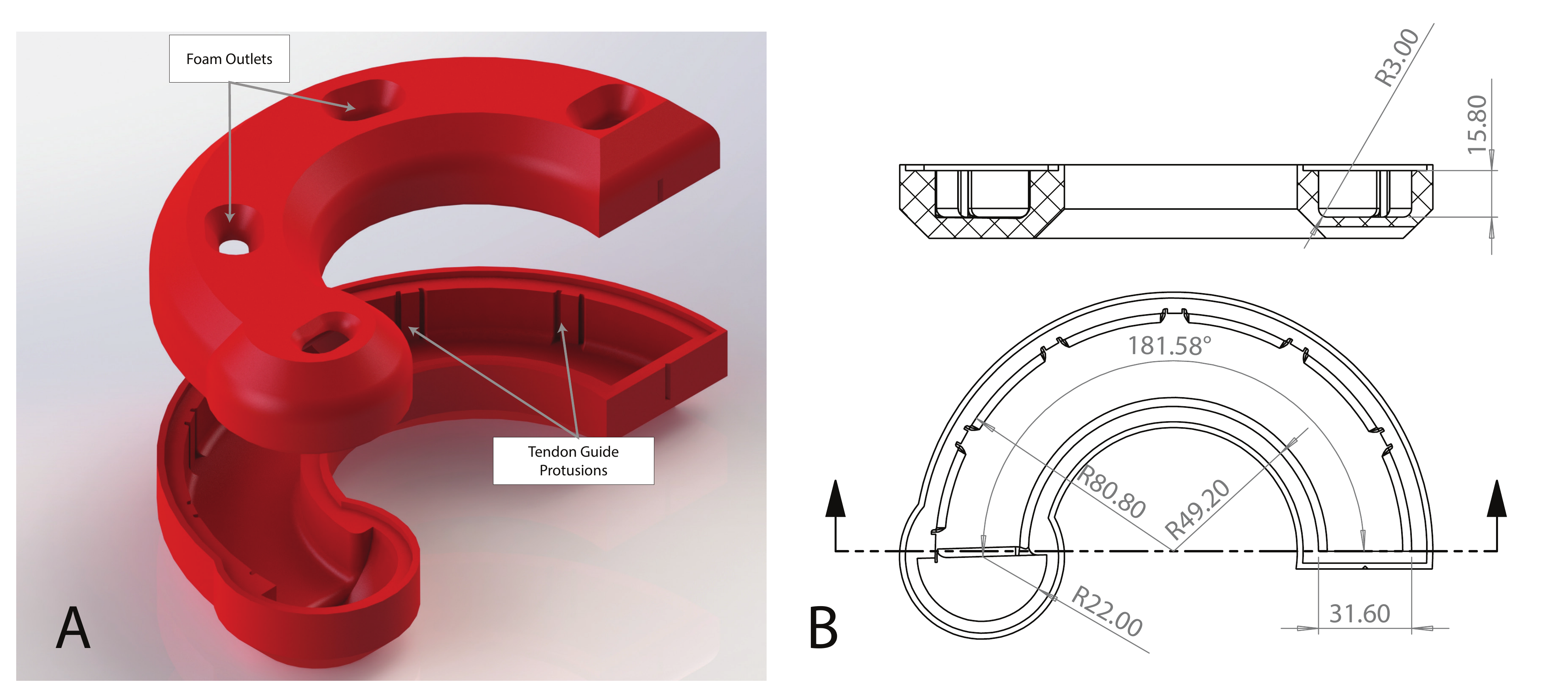}
     \caption{\textbf{Figure S4. Passively closed finger mold.} (\textbf{A}) The mold is 3D printed into a circular arc and consists of a top and bottom, such that the mold can be opened to remove the finger after it has cured. The top piece has outlets for excess liquid foam to escape during the curing process. Evenly spaced slits are placed in the interior of the mold; this leaves small foam protrusions on the cured finger that help mark the placement of the tendon guides. (\textbf{B}) Annotated drawing of the bottom mold, detailing the mold's radius and cross-sectional dimensions.}\label{fig:mold}
\end{figure*}

 \begin{figure*}[h]
    \centering
    \includegraphics[width=\textwidth]{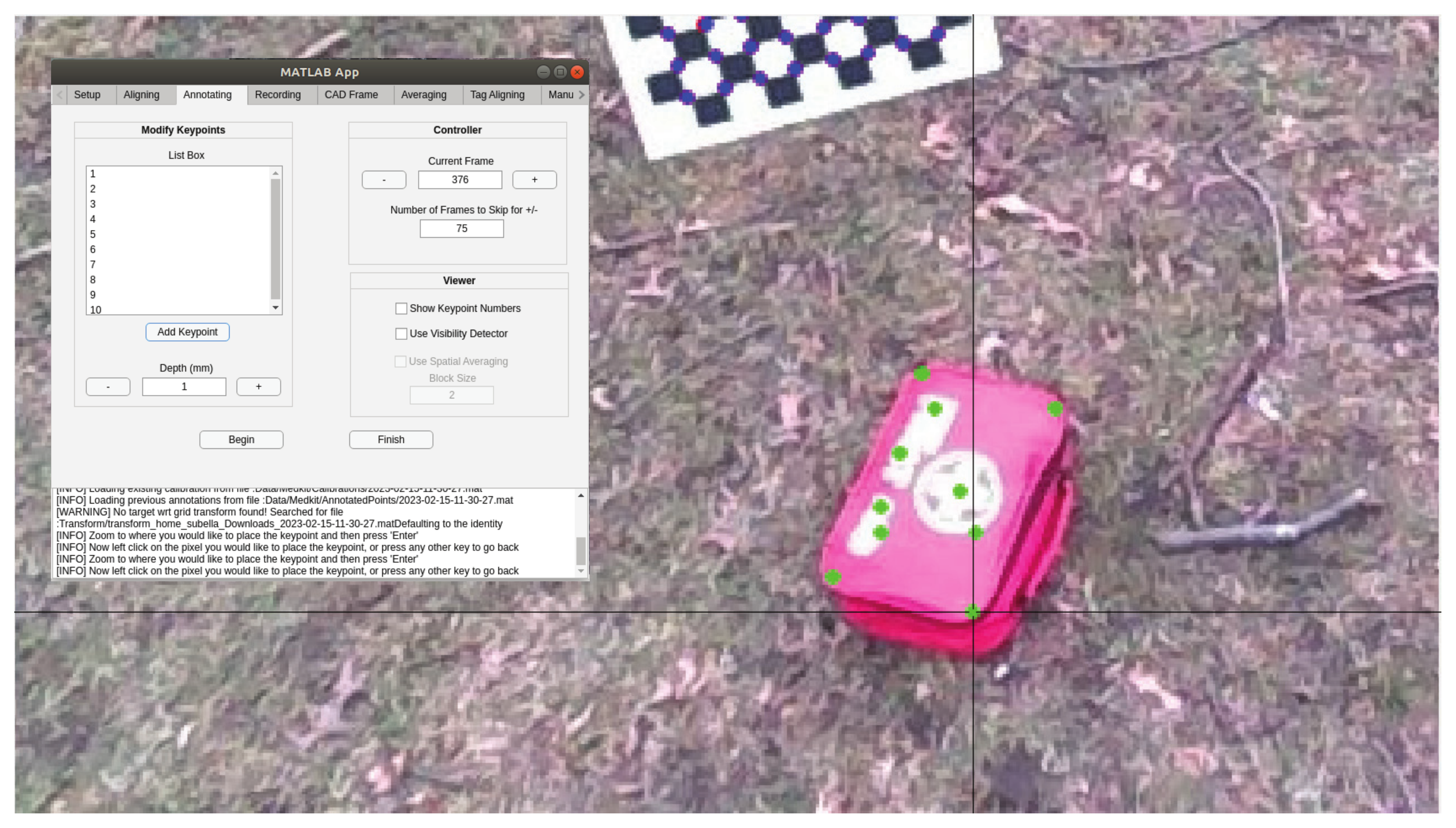}
     \caption{\textbf{Figure S5. Automated data annotation tool.} Example of the automated keypoint annotation Matlab GUI. The tool automatically parses RGB and depth images from a Rosbag and computes the camera pose for each image. Here, the user is manually placing a keypoint on the corner of the medkit. With known camera poses and leveraging the depth channel of the camera, this keypoint can then be projected back to pixel space for all subsequent frames. Once all the keypoints are labelled and refined, the user can switch to the \textit{recording} tab, which will annotate the pixel locations of keypoints in every image and format the data to be ready to train the network.}\label{fig:annotation}
\end{figure*}

\clearpage
 \begin{figure*}[h]
    \centering
    \includegraphics[width=\textwidth]{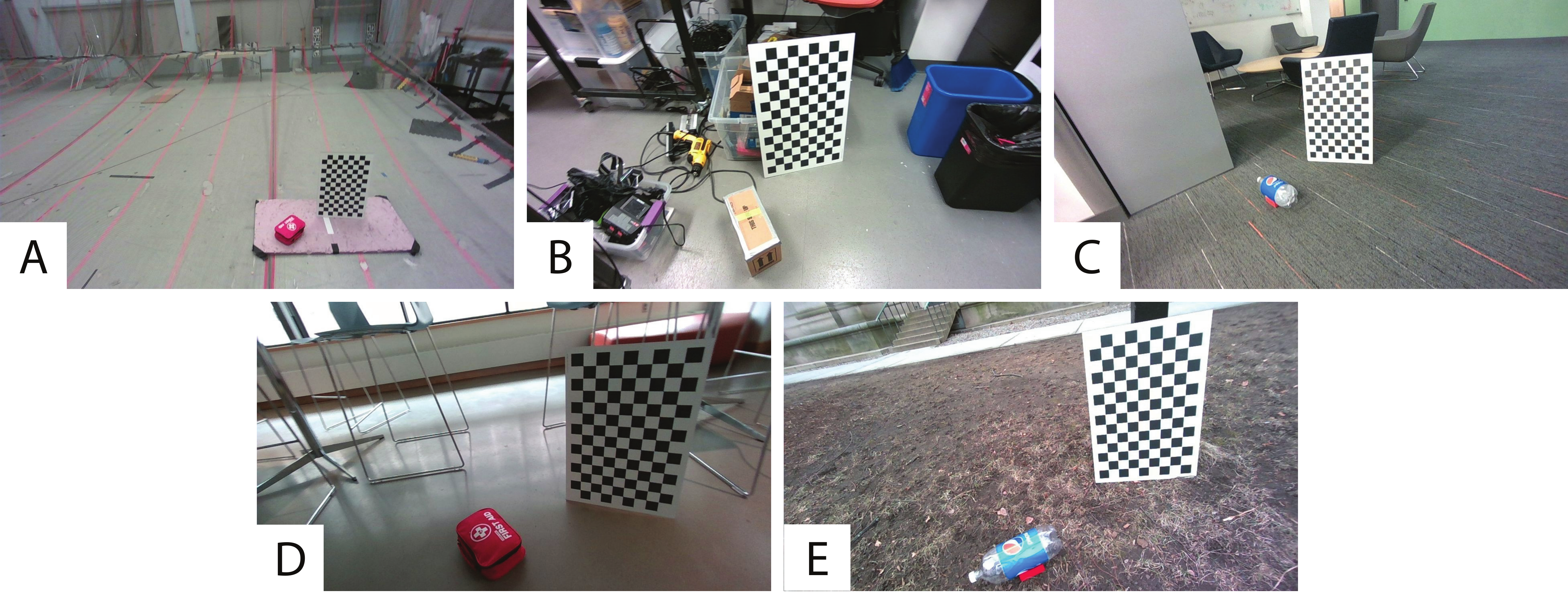}
     \caption{\textbf{Figure S6. Keypoint detector training environments.} To increase the generalizability of our keypoint detector, we collect training data in several environments: (\textbf{A}) motion capture room, (\textbf{B}) cluttered workshop, (\textbf{C}) lounge space, (\textbf{D}) foyer, and (\textbf{E}) outdoor field. We can quickly generate training data for new environments and targets with the aid of our automated data annotation tool.}\label{fig:environments}
\end{figure*}

 \begin{figure*}[h]
    \centering
    \includegraphics[width=0.5\textwidth]{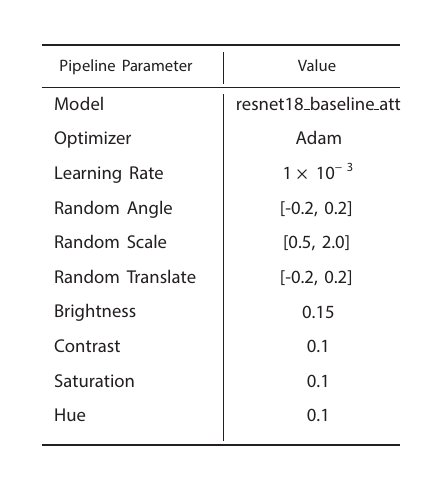}
     \caption{\textbf{Table S5. Keypoint detector training parameters.} All relevant, custom parameters used to train the model. The parameters correspond to values specified in trt-pose's training pipeline.}\label{fig:trt_params}
\end{figure*}

 \begin{figure*}[h]
    \centering
    \begin{equation}
    \label{eq:fixed_lag}
      \Omega_i^{-1} = 
      \begin{cases}
        \Sigma_{d} +
        \begin{bmatrix}
            t_{1} \; I_3 & 0 \\
            0 & r_{1} \; I_3 \\
        \end{bmatrix}
        &
        \\[15pt]
        \phantom{\Sigma_{d} \oplus \;}
        \begin{bmatrix}
            t_{i} \; I_3 & 0 \\
            0 & r_{i} \; I_3 \\
        \end{bmatrix}
        &
        \text{i=2,3,4}
  \end{cases}
  \tag{S1}
\end{equation}
    \includegraphics[width=0.8\textwidth]{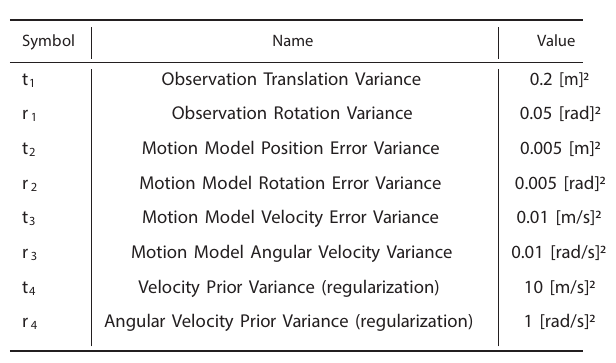}
     \caption{\textbf{Table S6. Fixed-lag smoother parameters.} Eq.~\ref{eq:fixed_lag} defines the weight matrices $\Omega_1,\hdots,\Omega_4$ in the fixed-lag smoother (Eq.~\ref{eq:fixed_lag_smoother}) based on the the parameters in this table. $\Omega_1$ scales the penalty on error between predicted and observed object pose, where $\Sigma_d$ is the covariance matrix of the drone's odometry estimate. Note that Eq.~\ref{eq:fixed_lag} assumes the measurement noise of the object relative to the camera is isotropic. In general, the observation noise must be rotated into the global frame. $\Omega_2$ scales the penalty on the object pose estimate differing from a constant velocity model. $\Omega_3$ scales the penalty associated with estimating higher changes in velocity (i.e., how constant we expect the velocity to be). $\Omega_4$ is a small regularization term reflecting a prior on the velocity and angular velocity not being too large. Note that our chosen values of $t_{4} = 10$~(m/s)\textsuperscript{2} and $r_{4}=1$~(rad/s)\textsuperscript{2} are quite reasonable, as any target we can hope to pick up will be moving at much slower speeds. This term is included to help ensure reliable initialization of the smoother.}\label{fig:fixed_lag}
\end{figure*}

\clearpage

\clearpage
\subsection*{Supplementary Text S1. Design Guidelines and Error Analysis}
We compute an approximate geometry-based bound on the error that can be tolerated for a successful grasp. The bound can be used to predict which objects are more likely to be grasped by a given gripper design and can inform future redesigns of the soft gripper for other applications.

Towards this goal, we measured the dimensions of the smallest enclosing 3D bounding box around  each object; the results are shown in Fig.~S3-A.
In the best case, we assume that any positioning error that still enables all four fingers to contact this bounding box could still result in a successful grasp, but positioning errors outside this range will likely fail. 
We can then approximate longitudinal, lateral, and vertical error bounds by using the gripper dimensions in its pre-grasp configuration and computing the maximum tolerable error (Fig.~S3-B-D) that still allows the four fingers to make contact. For our analysis, we assume that the gripper's center is aligned with the target's center laterally, and the top face of the target is vertically centered  between the lowest points on the rear and front fingers. 

For the longitudinal error bound (Fig.~S3-B), the solid green region depicts the approximate area the gripper encapsulates as it closes. Because the front fingers close in a circular arc, the effective longitudinal gripper length decreases as the fingers close. Given our assumed drone's vertical position, the effective longitudinal gripper length at the moment the front fingers would contact the target is $\delta_1=23.4$~cm. Our nominal control policy aims to have closed the gripper when the target is at the center of this longitudinal grasp length. If the gripper closes too early, the front fingers will contact the top of the target, rather than the side, and the grasp may fail. If the gripper closes too late, the gripper is more robust as the rear fingers can carry the target while the front fingers continue to close. Calling $\ell_1$ the longitudinal target length, the maximum tolerable longitudinal error can be computed as $\frac{\delta_1  - \ell_1}{2}$.

The lateral error bound (Fig.~S3-C) can be found similarly. If there exists so much lateral error the one half of the gripper does not make contact with the target, the grasp is unlikely to succeed. We measure the interior lateral gripper length between two pairs of fingers to be $\delta_2 = 9.8$ cm, and ---calling $\ell_2$ the lateral target length--- compute the lateral error bound as $\frac{\ell_2 - \delta_2}{2}$

For the vertical error bound (Fig.~S3-D), there exists two major failure modes: the rear fingers are too high to secure a grasp, or the front fingers are too low such that they collide with the target before grasp. We are less concerned with the rear fingers contacting the ground thanks to their compliance. We measure the vertical gripper length between the lowest point on the rear fingers and the lowest point on the front fingers as $\delta_3 = 15.2$ cm. As the top face of the target is assumed to be centered in this interval, the vertical error bound is simply $\delta_3/2$ for all targets. We note that in practice, we tune the vertical offset between the drone and the target at the grasp point empirically, rather than centering it.

We report these error bounds, as well as our worst-case, averaged observed errors in the longitudinal, lateral, and vertical axes in Tables S1-S3. To obtain our worst case total error for our experiments, we assume that the pose estimation error, VIO drift, and tracking error across each axis sum together in a way that maximizes error.

 In general, we see a bias in the pose estimation errors, particularly in the lateral axis. We conjecture that this bias is due to a slight misalignment between the vision and motion capture frames, and not an issue in our approach. Additionally, we note that our VIO metric is not a true reflection of the exact amount of drift at the time of grasp (see Section~\ref{sec:error_sources}). 
 Because the timing of the gripper closure only depends on the VIO state estimate and initial pose estimate of the target, the longitudinal tracking errors have a minimal effect on grasp performance, but are still reported in the total error metric. Finally, we do not have ground truth target pose data available for the two-liter 3 m/s experiment, and instead use the estimates from the two-liter 2 m/s experiment as there should be minimal difference.

We note that most experiments remained within our theoretical error bounds. We remark that these bounds are conservative: a grasp is unlikely to succeed outside of these bounds, but is not guaranteed to succeed inside, due to the complex interaction between target and soft gripper.

\subsection*{Supplementary Text S2. Evaluating Reference Polynomial in Object Frame}
Minimum-snap polynomials are widely used for generating quadrotor reference trajectories between waypoints.
As the system is differentially flat~\cite{Mellinger11icra,van1998real} in position and yaw, a reference trajectory will be feasible if it has a continuous 4th derivative (subject to actuator limits).
Each axis of the reference trajectory can be solved independently.
We solve for a separate polynomial for each position axis, resulting in $p_x(t), p_y(t), p_z(t)$, by solving the following optimization problem:
\begin{equation}
\begin{split}
\min_{p} &\int_{0}^{t_f} ||p^{4}(t)||^2 dt\\
&s.t.\\
    &\begin{aligned}
    p(0) &= x_0,   ~~~\dot{p}(0)&&=0,   ~~~&\ddot{p}(0) &= 0,\\
    p(t_g) &= x_g, ~~~\dot{p}(t_g)&&=v_g, ~~~&\ddot{p}(t_g) &= 0,\\
    p(t_f) &= x_f, ~~~\dot{p}(t_f)&&=0, ~~~&\ddot{p}(t_f) &= 0,
    \end{aligned}
\end{split}
\end{equation}
where $p$ is a 10th degree polynomial, $p^4(t)$ is its fourth derivative with respect to $t$, $x_0$ is the drone's current position in the relevant axis, $x_f$ is the desired post-grasp hover point, $t_g$ is the desired time of grasp, $x_g$ is the grasp waypoint, and $v_g$ is the grasp speed.
This optimization can be solved as a quadratic program (QP), although contrary to \cite{Mellinger11icra,richter2016polynomial} we solve for a single polynomial to represent both segments of the trajectory.
We use CVXOPT to solve the QP~\cite{cvxopt}.

The curve 
\begin{equation*}
\bm{x}(t) = [p_x(t), p_y(t), p_z(t)],~~~t \in [0, t_f]
\end{equation*}
represents the position setpoint over time of the drone during a grasp attempt.
Similarly,
\begin{align*}
\bm{v}(t) &= \bm{\dot{x}}(t) = [\dot{p}_x(t), \dot{p}_y(t), \dot{p}_z(t)],~~~t \in [0, t_f]\\
\bm{a}(t) &= \bm{\ddot{x}}(t) = [\ddot{p}_x(t), \ddot{p}_y(t), \ddot{p}_z(t)],~~~t \in [0, t_f].
\end{align*}
represent its velocity and acceleration.
The yaw of the drone is set such that the drone points toward its estimate of the object position at all times before grasp, and it maintains its yaw at the grasp point for times after $t_g$.

To enable the drone to grasp moving targets, we evaluate the grasp trajectory relative to the moving target frame.
As the lower-level controller expects a position, velocity, and acceleration setpoint in the drone's fixed odometry frame, we must transform $[\bm{x}(t),~\bm{v}(t),~\bm{a}(t)]$ from the object's moving frame to the fixed frame. We define the following variables:

\begin{table*}[h!]
\centering
\begin{tabularx}{0.8\textwidth}{l|p{3.5in}}
\toprule
Variable Names & Description\\
\midrule
$\bm{x}_o^f, \bm{v}_o^f, \bm{a}_o^f$ & Position, velocity, and acceleration setpoints for the drone in the fixed odometry frame, expressed in the basis of the fixed odometry frame. \\
\midrule
$\bm{x}, \bm{v}, \bm{a}$ & Position, velocity, and acceleration setpoints for the drone relative to the moving target frame, expressed in the moving object frame. \\
\midrule
$\bm{x}_o^m, \bm{v}_o^m, \bm{a}_o^m$ & Position, velocity, and acceleration setpoints for the drone relative to the moving target frame, expressed in the basis of the fixed odometry frame. For a rotation $\MR_m^f$ from the moving frame to the fixed frame, $[\bm{x}_o^m, \bm{v}_o^m, \bm{a}_o^m] = \MR_m^f [\bm{x}, \bm{v}, \bm{a}]$\\
\midrule
$\begin{aligned}&\bm{x}_m^f, \bm{v}_m^f, \bm{a}_m^f,\\ &\bm{\omega}_m^f, \bm{\alpha}_m^f\end{aligned}$ & 
\vspace{-5mm}Position, velocity, acceleration, angular velocity, and angular acceleration of the moving target frame relative to the fixed odometry frame, expressed in the fixed odometry frame.
\end{tabularx}
\end{table*}

The equations for transforming the setpoint into the global reference frame are then
\begin{align}
\bm{x}_o^f  &= \bm{x}_m^f + \bm{x}_o^m\\
\bm{v}_o^f  &= \bm{v}_m^f + \bm{\omega}_m^f \times \bm{x}_o^m + \bm{v}_o^m\\
\bm{a}_o^f  &= \bm{a}_m^f + \bm{\alpha}_m^f \times \bm{x}_o^m + \bm{\omega}_m^f \times \left[\bm{\omega}_m^f \times \bm{x}_o^m\right] + 2\bm{\omega}_m^f\times\bm{v}_o^m + \bm{a}_o^m.
\end{align}
See, \eg~\cite[Chapter~4]{kelly2013mobile}, for more exposition on this transformation. In our experiments, we assume the moving object's angular acceleration is zero, setting $\bm{\alpha}_m^f = \bm{0}$.

\clearpage

\end{document}